\documentclass[sn-basic]{sn-jnl}

\usepackage{graphicx}%
\usepackage{multirow}%
\usepackage{amsmath,amssymb,amsfonts}%
\usepackage{amsthm}%
\usepackage{mathrsfs}%
\usepackage[title]{appendix}%
\usepackage{xcolor}%
\usepackage{textcomp}%
\usepackage{manyfoot}%
\usepackage{booktabs}%
\usepackage{algpseudocode}%
\usepackage{listings}%

\usepackage{lmodern}
\usepackage[T1]{fontenc}

\usepackage[nice]{nicefrac}
\usepackage[ruled,vlined]{algorithm2e}
\usepackage[inline]{enumitem}
\usepackage{siunitx}
\usepackage{subcaption}
\usepackage{float}
\newcommand{\figref}[1]{Figure~\ref{#1}}
\newcommand{\algoref}[1]{Algorithm~\ref{#1}}
\newcommand{\aspref}[1]{Assumption~\ref{#1}}
\newcommand{\propref}[1]{Proposition~\ref{#1}}
\newcommand{\corref}[1]{Corollary~\ref{#1}}
\newcommand{\thmref}[1]{Theorem~\ref{#1}}
\newcommand{\tabref}[1]{Table~\ref{#1}}
\newcommand{\secref}[1]{Section~\ref{#1}}
\newcommand{\lmmref}[1]{Lemma~\ref{#1}}
\newcommand{\defref}[1]{Definition~\ref{#1}}

\DeclareMathOperator*{\argmin}{arg\,min}

\newcommand{\myedit}[1]{\textcolor{black}{#1}}
\usepackage{ulem}
\usepackage{makecell}

\newenvironment{editenv}
{\begingroup\color{black}}
{\endgroup}

\usepackage{pifont}
\newcommand{\cmark}{\text{\ding{51}}}
\newcommand{\xmark}{\text{\ding{56}}}

\usepackage{threeparttable} 
\usepackage{booktabs}

\theoremstyle{thmstyleone}%
\newtheorem{theorem}{Theorem}[section]
\newtheorem{proposition}{Proposition}[section]%
\newtheorem{assumption}{Assumption}[section]
\newtheorem{corollary}{Corollary}[section]
\newtheorem{lemma}{Lemma}[section]
\newtheorem{definition}{Definition}[section]

\begin{document}

\title[Test]{Stochastic Online Optimization for Cyber-Physical and Robotic Systems}


\author*[1,2]{\fnm{Hao} \sur{Ma}}\email{hao.ma@tuebingen.mpg.de}

\author[2]{\fnm{Melanie} \sur{Zeilinger}}\email{mzeilinger@ethz.ch}

\author[1]{\fnm{Michael} \sur{Muehlebach}}\email{michael.muehlebach@tuebingen.mpg.de}

\affil*[1]{\orgdiv{Learning and Dynamical Systems Group}, \orgname{Max Planck Institute for Intelligent Systems}, \orgaddress{\street{Max-Planck-Ring 4}, \city{T\"ubingen}, \postcode{72076}, \state{Baden-W\"urttemberg}, \country{Germany}}}

\affil[2]{\orgdiv{Institute for Dynamic Systems and Control}, \orgname{ETH Z\"urich}, \orgaddress{\street{Leonhardstrasse 21}, \city{Z\"urich}, \postcode{8092}, \state{Z\"urich}, \country{Switzerland}}}


\abstract{We propose a gradient-based online optimization framework for solving stochastic programming problems that frequently arise in the context of cyber-physical and robotic systems. Our problem formulation accommodates constraints that model the evolution of a cyber-physical system, which has, in general, a continuous state and action space, is nonlinear, and where the state is only partially observed. We also incorporate an approximate model of the dynamics as prior knowledge into the learning process and show that even rough estimates of the dynamics can significantly improve the convergence of our rithms. Our online optimization framework encompasses both gradient descent and quasi-Newton methods, and we provide a unified convergence analysis of our algorithms in a non-convex setting. We also characterize the impact of modeling errors in the system dynamics on the convergence rate of the algorithms. Finally, we evaluate our algorithms in simulations of a flexible beam, a four-legged walking robot, and in real-world experiments with a ping-pong playing robot.}




\maketitle

\section{Introduction}
\label{sec:introduction}
The increasing availability of sensors across various domains has led to the generation of vast volumes of data, ideal for analysis and training. However, a significant challenge arises from the traditional ``Sampling-Training-Deployment'' mode of machine learning algorithms. Once trained, most models remain static during deployment, unable to benefit from the continuous influx of new data. This limitation means that models risk becoming outdated as the environment evolves and new information emerges, leaving a substantial amount of potentially valuable data unused. This not only hinders improvements in performance but also falls short of enabling systems to continuously learn, adapt, and improve. Moreover, retraining models from scratch with new data is neither an economical nor a long-term solution. This issue is particularly prevalent in robotics, where deployed models struggle to adapt to ever-changing environments and continuous streams of new information and data.

\begin{editenv}
\subsection{Motivation}
\label{sec:motivation}
\citet{hazanIntroductionOnlineControl2025} explore the challenge of controlling a drone to navigate from a source to a destination under unpredictable environmental conditions, such as weather and terrain. While indoor environments can be effectively managed using existing technologies, outdoor conditions present a significantly more complex problem, often requiring online adaptation. Not only does the environment in which a cyber-physical system operates constantly change (e.g. due to adverse weather conditions), but the system dynamics also drift over time. Factors such as battery degradation and mechanical wear gradually lead to performance deterioratation. Thus, the conventional static ``Sampling-Training-Deployment'' paradigm in machine learning, which lacks online adaptivity, may severely limit performance of cyber-physical systems in complex environments.


A similar challenge is evident in reinforcement learning (RL). RL methods suffer from sample-inefficiency~\citep{laskinReinforcementLearningAugmented2020,heessEmergenceLocomotionBehaviours2017,kalashnikovQTOptScalableDeep2018} and are significantly impacted by the sim-to-real gap~\citep{shahAirSimHighFidelityVisual2017,dosovitskiyCARLAOpenUrban2017,furrerRotorSModularGazebo2016,mccordDistributedProgressiveFormation2019}. Even for a static environment, trained models often require extensive fine-tuning before they can be successfully deployed in real-world applications, which requires extensive trial-and-error and human supervision, and might cause damage and unexpected harm.

To address these challenges, we propose a shift toward a new learning paradigm that enables cyber-physical systems to continuously learn from streaming data, allowing them to adapt to changing environments and non-stationary system dynamics in real time. This approach leverages online learning, a machine learning framework designed to efficiently incorporate data streams and update models dynamically, to overcome the limitations of the conventional ``Sampling-Training-Deployment'' paradigm enabling continuous performance improvement in complex real-world settings.
\end{editenv}

\subsection{Online Learning}
\label{sec:online_learning}
Online learning aims to minimize the expected regret as follows~\citep{bubeckRegretAnalysisStochastic2012,neuExploreNoMore}:
\begin{equation}
    \mathrm{Regret} \left(\mathcal{A}\right) = \mathbb{E} \left[ \sum_{t=1}^{T} f \left(\omega_t; \zeta_t\right) \right]- \min_{\omega \in \Omega} 
    \mathbb{E}  \left[ \sum_{t=1}^{T} f \left(\omega; \zeta_t \right) \right],~\zeta_t \stackrel{\text{i.i.d.}}{\sim} \mathcal{D}, 
    \label{eq:stochastic_setting}
\end{equation}
where $\mathcal{A}$ denotes the online optimization algorithm that is used for performance optimization and $f$ are stochastic loss functions, where the random variable $\zeta_t$ is independently and identically sampled from the unknown distribution $\mathcal{D}$. The expectation is taken with respect to the loss functions and the decision variables $\omega_t \in \Omega$. The decision variables $\omega_t$ are generated by the algorithm $\mathcal{A}$ from a closed and convex set $\Omega \subseteq \mathbb{R}^{n_{\omega}}$, where $n_{\omega} \in \mathbb{N}_{+}$ denotes the number of decision variables $\omega$. The loss functions $f$ are typically assumed to be convex and bounded~\citep{hazanIntroductionOnlineConvex2021a, shalev-shwartzOnlineLearningOnline2012, hallOnlineConvexOptimization2015b}. However, one of the primary objectives of this article is to bridge the gap between theory and practice, enabling online learning algorithms to be deployed on cyber-physical systems. Therefore, in this article, we abandon the convexity assumption and instead rely on a more general smoothness assumption. In addition, $T$ denotes the total number of iterations. Intuitively, we claim that an online optimization algorithm $\mathcal{A}$ performs well if the regret induced by this algorithm is sub-linear as a function of $T$ (i.e, $\textrm{Regret}\left(\mathcal{A}\right)=o\left(T\right)$), since this implies that on average, the algorithm performs as well as the best fixed decision variables $\omega^{\star}=\arg \min_{\omega\in\Omega} \mathbb{E} \left[\sum_{t=1}^{T} f \left(\omega; \zeta_t \right)\right] $ in hindsight. Therefore, when $T$ is large enough or tends to infinity, online learning can cope with continually growing streams of data at the algorithm level, so that the decision variables $\omega_t$ continuously improve performance.

\citet{zinkevichOnlineConvexProgramming} was one of the first researchers to realize that simple gradient-descent strategies are effective at solving~\eqref{eq:stochastic_setting}. Since then, a plethora of online learning algorithms have emerged and their effectiveness has been demonstrated in practical settings. These algorithms can be broadly categorized into two groups: The first group comprises algorithms that are targeted at linear loss functions, also referred to as the prediction with expert advice problem~\citep{hazanIntroductionOnlineConvex2021a,abernethyCompetingDarkEfficient,shalev-shwartzPrimaldualPerspectiveOnline2007}, with notable examples being the Follow the Leader (FTL) and Follow the Regularized Leader (FTRL) algorithms, as well as the hedge algorithm~\citep{freundDecisiontheoreticGeneralizationOnline1997, littlestoneWeightedMajorityAlgorithm1994,muehlebachAdaptiveDecisionMakingConstraints2023}, and the Perceptron~\citep{novikoff1962convergence}. Other online optimization algorithms have also been proposed to tackle the problem of prediction with expert advice, such as the exponentially weighted average forecaster and the Greedy Forecaster~\citep{cesa-bianchiPredictionLearningGames2006}, as well as the upper confidence bound algorithm~\citep{auerFinitetimeAnalysisMultiarmed2002}. Compared to these online learning methods that are targeted to linear loss functions, gradient-based methods can handle more general loss functions and sometimes even achieve faster convergence rates~\citep{ruderOverviewGradientDescent2017a}. In this article, we will focus on gradient-based online optimization and learning algorithms. Gradients provide local information of the loss functions, which, compared to gradient-free algorithms, is known to expedite the convergence process~\citep{nesterovRandomGradientFreeMinimization2017}. However, in the stochastic programming problems that are considered herein, gradient information is often challenging or even impossible to obtain, which is why many RL algorithms rely, at their core, on zeroth-order optimization. In contrast, we advocate with our work the use of approximate, model-based gradients and demonstrate that even rough estimates are enough to achieve convergence. We also quantitatively characterize the impact of estimation errors on convergence and highlight that in contrast to zeroth-order methods, our rates are dimension independent.

Gradient-based online optimization algorithms can be further categorized based on whether higher-order derivatives are used. Examples of first-order gradient-based algorithms include the Passive-Aggressive Online Learning~\citep{crammerOnlinePassiveAggressiveAlgorithms}, the approximate large margin agorithm~\citep{gentileNewApproximateMaximal2000}, the Online Mirror Descent (OMD)~\citep{hazanIntroductionOnlineConvex2021a,bubeck2011introduction}, and the relaxed online maximum margin algorithm~\citep{liRelaxedOnlineMaximum1999}. Among these, the most widely employed algorithm is Online Gradient Descent (OGD)~\citep{dekelOptimalDistributedOnline, hazanAdaptiveOnlineGradient2007, hazanLogarithmicRegretAlgorithms2007a,NEURIPS2023_a6f27630}, which is summarized in \algoref{algo:online_gradient_descent}.
\SetKwInOut{Given}{Given}
\SetKwInOut{Input}{Input}
\SetKwInOut{Return}{Return}
\begin{algorithm}[H]
\SetAlgoLined
\Input{initial parameters $\omega_1 \in \Omega$, iterations $T$, step length $\left\{\eta_t\right\}_{t=1}^{T}$}
\For{$t=1$ to $T$}
{
{\sc Implementation}: play $\omega_{t}$ and observe cost $f_{t}\left(\omega_{t}\right)$\;
{\sc Update}: $\widehat{\omega}_{t+1} = \omega_{t} - \eta_{t} \nabla f_{t} \left(\omega_{t}\right)$\;
{\sc Projection}: $\omega_{t+1} = \Pi_{\Omega} \left(\widehat{\omega}_{t+1}\right)$\;
}
\Return{$\omega_{T+1}$}
\caption{Online Gradient Descent}
\label{algo:online_gradient_descent}
\end{algorithm}
We note that $\Pi_{\Omega} \left(\cdot\right)$ denotes the projection to the feasible set $\Omega$, and is defined as follows:
\begin{equation}
    \omega_t = \Pi_{\Omega}\left(\widehat{\omega}_t\right) = \argmin_{\omega \in \Omega} \left|\omega - \widehat{\omega}_t \right|^2,
    \label{eq:projection}
\end{equation}
where $\left|\cdot\right|$ denotes the $\ell_2$-norm.

Sometimes, even faster convergence rates can be obtained with second-order methods, which not only require gradient information, but also the Hessian~\citep{boydConvexOptimization2004b,nocedalNumericalOptimization2006a}. Examples include the second-order Perceptron~\citep{cesa-bianchiSecondOrderPerceptronAlgorithm2005}, the Confidence Weighted Learning~\citep{dredzeConfidenceweightedLinearClassification2008}, the Adaptive Regularization of Weight Vectors~\citep{crammerAdaptiveRegularizationWeight2009}, and the Online Newton Step (ONS)~\citep{hazanIntroductionOnlineConvex2021a}. However, in practical applications related to the control of cyber-physical systems, gradients are already difficult to obtain, let alone second-order information. In addition, second-order methods are plagued by the following shortcomings~\citep{martens2015optimizing, reddiConvergenceAdam2019, bottouLargeScaleMachineLearning2010,lecunEfficientBackProp1998,agarwal2017second}: 
\begin{enumerate*}
\item [i)] the methods have a high computational complexity per iteration as they require assembling and inverting the Hessian, which is problematic for high-dimensional problems; \item[ii)] the methods can be sensitive to noise (inexact gradient and Hessian evaluation), which can result in oscillations or divergence.
\end{enumerate*} Thus, in this article, we introduce an efficient approximation method for the Hessian matrix based on a trust-region method. Additionally, we provide a quantitative analysis of the sensitivity of our proposed second-order method to gradient estimation errors. \begin{editenv}A selection of the most important online learning algorithms, combined with their underlying assumptions and corresponding results, is summarized in \tabref{tab:oco-summary}.\end{editenv}
\begin{table}[t]
\centering
\caption{Comparison of representative algorithms for online convex optimization, including their assumptions, regret bounds, and corresponding lower bounds.}
\label{tab:oco-summary}
\begin{tabular}{lccc}
\toprule
\textbf{Algorithm} 
& \textbf{Assumptions} 
& \textbf{Regret Bound} 
& \textbf{Lower Bound} \\
\midrule
FTRL 
& \makecell{convex losses\\ strongly convex regularizer}
& $\mathcal{O}(\ln T)$ 
& $\Omega(\sqrt{T})$ \\

FTRL 
& \makecell{strongly convex losses\\ strongly convex regularizer} 
& $\mathcal{O}(\ln T)$ 
& $\Omega(\ln T)$ \\

OGD 
& \makecell{convex losses\\ bounded domain}
& $\mathcal{O}(\sqrt{T})$ 
& $\Omega(\sqrt{T})$ \\

OGD 
& \makecell{strongly convex losses\\ bounded domain}
& $\mathcal{O}(\ln T)$ 
& $\Omega(\ln T)$ \\

OMD 
&  \makecell{convex losses\\ strongly convex mirror map} 
& $\mathcal{O}(\sqrt{T})$ 
& $\Omega(\sqrt{T})$ \\

ONS 
&  \makecell{exp-concave or\\ strongly convex losses} 
& $\mathcal{O}(\ln T)$ 
& $\Omega(\ln T)$ \\
\bottomrule
\end{tabular}
\vspace{0.5em}
\footnotesize
\textbf{Abbreviations:} 
FTRL: Follow-The-Regularized-Leader; 
OGD: Online Gradient Descent; 
OMD: Online Mirror Descent; 
ONS: Online Newton Step.
\end{table}

\subsection{Related Work}
\label{sec:related_work}
\begin{editenv}
Online learning in cyber-physical systems and robotics, often referred to as online control, has long remained an underexplored area. Most of the existing work is confined to linear systems, typically in a setting where the dynamics of the system (and environment) are linear in both state and action, perturbed by Gaussian noise, and the loss function is quadratic in state and action, which corresponds to the well-known linear quadratic regulator (LQR) problem in control theory. When the system is unknown, researchers usually employ online learning algorithms to estimate the system parameters, then use tools from robust control theory to solve the LQR problem, thereby obtaining a linear feedback controller. For instance, \citet{abbasi-yadkoriRegretBoundsAdaptive2011} use online parameter estimation to achieve a control policy whose regret bound is $\mathcal{O}\left(T^{\nicefrac{1}{2}}\right)$. Under the same setting, \citet{deanRegretBoundsRobust2018} have introduced a computationally efficient alternative that attains a regret bound of $\mathcal{O}\left(T^{\nicefrac{2}{3}}\right)$, while also pointing out that achieving this rate requires the estimation errors of the system parameters to converge to a certain threshold. Moreover, \citet{cohenLearningLinearQuadraticRegulators2019} reformulate the LQR problem as a convex semi-definite program, improving the regret bound to $\mathcal{O}\left(T^{\nicefrac{1}{2}}\right)$ while preserving computational efficiency, and the regret bound is growing cubically with the system dimension. \citet{maniaCertaintyEquivalenceEfficient2019} further relax the assumptions by considering that the system state is only partially observable, a setting with significant practical implications that we also adopt in our work, resulting in a regret bound nearly at $\mathcal{O}\left(T^{\nicefrac{1}{2}}\right)$. The regret bound explicitly depends on the error between the approximate system and the true system, and achieving the proposed regret requires that this error is sufficiently small. All these approaches share the common paradigm of first estimating the system parameters online and then using control-theoretic tools to solve the LQR problem to obtain a linear feedback control policy.

In addition, \citet{abbasi-yadkoriModelFreeLinearQuadratic2018} have proposed a model-free algorithm for controlling linear unknown systems subject to quadratic loss functions with theoretical guarantees. Numerical experiments demonstrate that this method dramatically outperforms standard policy iteration, although its performance is still inferior to that of a model-based approach, with a regret bound of $\mathcal{O}\left(T^{\nicefrac{2}{3} + \xi}\right)$ for any small $\xi > 0$, provided that the time horizon satisfies $T > C^{\nicefrac{1}{\xi}}$ for some constant $C$. Moreover, the regret bound explicitly depends on the system dimension.

Various studies have explored alternative settings within the framework of linear systems. For example, a series of works assume that the loss function is adversarially chosen: \citet{abbasi-yadkoriTrackingAdversarialTargets} assume that the loss function remains quadratic but allows adversarially chosen tracking targets, with the system operating without noise. Since the system parameters are known, the proposed algorithm directly learns the control policy, achieving a regret bound of $\mathcal{O}\large(\left(\log T\right)^2\large)$. Moreover, when the system transition dynamics are also adversarially chosen, the regret bound becomes $\mathcal{O}\left(T^{\nicefrac{1}{2}}\right)$. In this setting, the regret bound additionally scales logarithmically with the size of the set of linear policies. \citet{cohenOnlineLinearQuadratic2018} consider a linear time-invariant system with known (stochastic) dynamics, where the loss function remains quadratic but the parameters can be adversarially chosen. In this case, a gradient-based algorithm is proposed to directly learn the control policy online, yielding a regret bound of $\mathcal{O}\left(T^{\nicefrac{1}{2}}\right)$. Moreover, the regret bound scales sublinearly with the system dimension. \citet{cohenOnlineLinearQuadratic2018} provide simulation experiments evaluating this result where real-world data is used for the system dynamics while varying the costs and running the algorithms in simulation.

Another line of research has studied the convergence of different algorithms under adversarial settings, where the process noise is no longer stochastic but is instead composed of bounded adversarial disturbances, and any adversarial convex loss function is allowed. In this setting, traditional worst-case approaches in the control literature, such as $\mathcal{H}_{\infty}$ control and its variants, tend to be overly pessimistic. For a known linear system, \citet{agarwalOnlineControlAdversarial2019} employ a gradient-based algorithm to predict the weight matrix of the disturbance, thereby indirectly parameterizing the control input and achieving a regret bound of $\mathcal{O}\left(T^{\nicefrac{1}{2}}\right)$; \citet{hazanNonstochasticControlProblem2020} extend this result to unknown systems, attaining a regret bound of $\mathcal{O}\left(T^{\nicefrac{2}{3}}\right)$; and \citet{yanOnlineNonstochasticControl} go even further by assuming that the system state is partially observable and use a gradient-free method, which results in a regret bound of $\mathcal{O}\left(T^{\nicefrac{3}{4}}\right)$. Their algorithm is evaluated in both synthetic linear and simulated nonlinear environments. It is worth noting that the regret bounds proposed in the aforementioned works all exhibit an explicit dependence on the system dimension.

In practice, however, most, if not all, systems are nonlinear. To address this, \citet{graduAdaptiveRegretControl2022} linearize a nonlinear system locally, transforming it into a time-varying linear system with bounded adversarial disturbances and adversarial convex loss functions. At the same time, \citet{graduAdaptiveRegretControl2022} employ the concept of adaptive regret to handle changing environments. The algorithm implicitly tracks the locally optimal policy as the dynamics of the linear time-variant system change, and the adaptive regret achieves the bound of $\mathcal{O}\big(\mathrm{OPT}^{\nicefrac{1}{2}}\big)$, where OPT denotes the cost of the best policy in hindsight over the entire horizon. 

The paradigm of iterative system identification and control design has been studied in the adaptive control community for a long time, and various shortcomings have been pointed out~\citep{andersonFailuresAdaptiveControl2005}. An important shortcoming is that the identified system dynamics are only accurate for the given feedback controller that is applied in the identification step, and may not accurately describe the closed-loop dynamics when a different controller is applied, which may lead to unexpected results and failures. To address such scenarios with large model-uncertainty, \citet{hoOnlineRobustControl2021} propose an alternative pipeline that uses only an approximate system dynamics and learns to control online. This idea aligns with our work, and we further quantitatively characterize the upper bound on system uncertainty allowed and its effect on convergence rates. Unlike traditional online control methods, \citet{hoOnlineRobustControl2021} consider binary loss functions, where the cost is one when the system state and action do not meet a specified condition (indicating a mistake) and zero otherwise. The proposed algorithm is proven to incur only a finite number of mistakes in the long run, and simulation experiments (e.g., cart-pole swing-up) confirm this result. Furthermore, \citet{linOnlinePolicyOptimization2024} introduce a meta-framework that simultaneously employs an online estimator for system parameters and online policy optimization for the controller. For a time-varying nonlinear system with Gaussian noise, this framework is shown to achieve a local regret bound of $\mathcal{O}\left(T^{\nicefrac{5}{6}}\right)$. Here, the local regret is defined as the sum of the norms of the partial derivatives of the loss function with respect to the state for each iteration; \citet{linOnlinePolicyOptimization2024} argue that in nonconvex settings, this performance metric is more meaningful than traditional regret, and we adopt this metric in our work to characterize algorithm convergence. A detailed comparison of our work and related results is summarized in~\tabref{tab:online_control_linear_systems}.
\begin{sidewaystable}
    \centering
    \begin{minipage}{\textwidth}
    \caption{Comparisons of our work and previous ones regarding problem setups and results.}
    \label{tab:online_control_linear_systems}
    \begin{tabular}{c|cccccccc}  
        \toprule
        \textbf{References} 
        & \makecell{\textbf{Unknown} \\ \textbf{System}} 
        & \makecell{\textbf{Loss} \\ \textbf{Function}} 
        & \textbf{Disturbance}
        & \makecell{\textbf{Partial}\\ \textbf{Feedback}}  
        & \makecell{\textbf{Modeling} \\ \textbf{Error} \\ \textbf{Analysis}}  
        & \makecell{\textbf{Evaluation} \\ \textbf{in Simulation}}  
        & \makecell{\textbf{Evaluation in} \\ \textbf{Real World}}  
        & \textbf{Regret} \\        
        \midrule
        \multicolumn{9}{c}{\textbf{Methods for Linear Systems}} \\
        \midrule
        \citet{abbasi-yadkoriRegretBoundsAdaptive2011}
        & $\cmark$ & quadratic & Gaussian & $\xmark$ & $\xmark$ & $\xmark$ & $\xmark$ & $\mathcal{O}\left(T^{\nicefrac{1}{2}}\right)$\\
        \citet{deanRegretBoundsRobust2018} 
        & $\cmark$ & quadratic & Gaussian & $\xmark$ & $\xmark$ & $\cmark$  & $\xmark$ & $\mathcal{O}\left(T^{\nicefrac{2}{3}}\right)$\\
        \citet{cohenLearningLinearQuadraticRegulators2019} 
        & $\cmark$ & quadratic & Gaussian & $\xmark$ & $\xmark$ & $\xmark$ & $\xmark$ & $\mathcal{O}\left(T^{\nicefrac{1}{2}}\right)$\\
        \citet{maniaCertaintyEquivalenceEfficient2019} 
        & $\cmark$ & quadratic & Gaussian & $\cmark$ & $\cmark$ & $\xmark$ & $\xmark$ & $\mathcal{O}\left(T^{\nicefrac{1}{2}}\right)$\\
        \citet{abbasi-yadkoriModelFreeLinearQuadratic2018} 
        & $\cmark$ & quadratic & Gaussian & $\xmark$ & $\xmark$ & $\cmark$ & $\xmark$ & $\mathcal{O}\left(T^{\nicefrac{2}{3} + \xi}\right)$\footnote{For any small $\xi > 0$ and time horizon satisfies $T > C^{\nicefrac{1}{\xi}}$ for a constant $C$.}\\
        \citet{abbasi-yadkoriTrackingAdversarialTargets} 
        & $\xmark$ & \makecell{quadratic \\ adversarial} & $\xmark$ & $\xmark$ & $\xmark$ & $\xmark$ & $\xmark$ & $\mathcal{O}\left(\left(\log T\right)^2\right)$\\
        \citet{abbasi-yadkoriTrackingAdversarialTargets} 
        & $\cmark$ & \makecell{quadratic \\ adversarial} & $\xmark$ & $\xmark$ & $\xmark$ & $\xmark$ & $\xmark$ & $\mathcal{O}\left(T^{\nicefrac{1}{2}}\right)$\\
        \citet{cohenOnlineLinearQuadratic2018} 
        & $\xmark$ & \makecell{quadratic \\ adversarial} & \makecell{known \\ dynamics} & $\xmark$ & $\xmark$ & $\cmark$ & $\xmark$ & $\mathcal{O}\left(T^{\nicefrac{1}{2}}\right)$\\
        \citet{agarwalOnlineControlAdversarial2019} 
        & $\xmark$ & \makecell{convex \\ adversarial} & adversarial & $\xmark$ & $\xmark$ & $\xmark$ & $\xmark$ &$\mathcal{O}\left(T^{\nicefrac{1}{2}}\right)$\\
        \citet{hazanNonstochasticControlProblem2020} 
        & $\cmark$ & \makecell{convex \\ adversarial} & adversarial & $\xmark$ & $\xmark$ & $\xmark$ & $\xmark$ &$\mathcal{O}\left(T^{\nicefrac{2}{3}}\right)$\\
        \citet{yanOnlineNonstochasticControl} 
        & $\cmark$ & \makecell{convex \\ adversarial} & adversarial & $\cmark$  & $\xmark$  & $\cmark$ & $\xmark$ &$\mathcal{O}\left(T^{\nicefrac{3}{4}}\right)$\\
        \midrule
        \multicolumn{9}{c}{\textbf{Methods for Nonlinear Systems}} \\
        \midrule
        \citet{graduAdaptiveRegretControl2022} 
        & $\cmark$ &  \makecell{convex \\ adversarial} & adversarial & $\xmark$ & $\xmark$ & $\xmark$ & $\xmark$ &$\mathcal{O}\left(\mathrm{OPT}^{\nicefrac{1}{2}}\right)$\footnote{OPT is the cost of the best policy in hindsight over the entire horizon.}\\
        \citet{hoOnlineRobustControl2021} 
        & $\cmark$ & binary & $\xmark$ & $\xmark$ & $\cmark$ & $\cmark$ & $\xmark$ &\makecell{finite mistake \\ bound}\footnote{The convergence rate is characterized by the adaptive regret bound.}\\
        \citet{linOnlinePolicyOptimization2024} 
        & $\cmark$ & \makecell{nonconvex \\ adversarial} & Gaussian & $\xmark$ & $\xmark$ & $\cmark$ & $\xmark$ &$\mathcal{O}\left(T^{\nicefrac{5}{6}}\right)$\footnote{The convergence rate is characterized by the local regret bound.}\\
        \midrule
        \textbf{Ours} 
        & $\cmark$ & \makecell{nonconvex \\ stochastic} & stochastic & $\cmark$ & $\cmark$ & $\cmark$ & $\cmark$ &$\mathcal{O}\left(T^{\nicefrac{1}{2}}\right)$\footnote{The convergence rate is characterized by the local regret bound.}\\
        \bottomrule
    \end{tabular}
    \end{minipage}
\end{sidewaystable}

Despite the thorough and solid exploration of online control in previous studies, evaluations remain limited to simple numerical experiments. In robotics, online learning is often used for high-level planning tasks, rarely involving control policy learning~\citep{bellicosoDynamicLocomotionOnline2018,wangFastOnlineOptimization2022,chignoliOnlineTrajectoryOptimization2021}. This creates a significant gap between theory and practice. Our work bridges this gap by proposing a solution for online control of nonlinear systems, supported by real-world experiments. To the best of our knowledge, our work is pioneering in this regard.
\end{editenv}

\subsection{Contribution}
\begin{editenv}
    As we mentioned in \secref{sec:online_learning}, gradient‐based methods enjoy faster convergence rates compared to gradient-free approaches, and this rate further improves with the use of higher-order derivatives, a point that \citet{abbasi-yadkoriModelFreeLinearQuadratic2018,agarwalOnlineControlAdversarial2019,hazanNonstochasticControlProblem2020} confirms. However, very few studies in online control have applied gradient-based online learning in practice. We believe this is because, unlike in tasks such as text or image classification, obtaining gradients in the context of cyber-physical systems is particularly challenging. On the one hand, the system dynamics are often complex; on the other hand, the interactions with environments may be unknown, change over time, or difficult to model. In addition, some of our common assumptions in online learning settings tend to be ineffective, such as convexity of the loss functions, which often renders the theoretical guarantees vacuous. This explains why, despite its robust theoretical foundations, online control has rarely been implemented in real-world cyber-physical systems. 

    The purpose of this article is to bridge the gap between theory and practice in online control by proposing a framework that enables gradient‐based online optimization algorithms to be directly applied to cyber‐physical systems. Unlike most online control approaches, which typically perform online system identification followed by solving programming problems to derive control policies, we follow the idea of \citet{hoOnlineRobustControl2021}: we use an approximate model of the system and learn the control policy online. In contrast to \citet{heGrayBoxNonlinearFeedback2024,chanRobustFeedbackOptimization2025}, which concentrate on feedback optimization in the online learning, our approach is applicable to both open-loop and closed-loop optimization settings. Our approach tolerates significant uncertainty in the approximate model, which is denoted by the modeling error modulus $\kappa$. We not only theoretically characterize an upper bound on $\kappa$ and quantitatively assess its impact on convergence rates, but we also evaluate our theory through a variety of approximate models. Finally, building on the efficiency of quasi‐Newton methods, under appropriate assumptions, our regret bound improves to $\mathcal{O}\left(T^{\nicefrac{1}{2}}\right)$. In addition, compared to the works of \citet{cohenLearningLinearQuadraticRegulators2019,abbasi-yadkoriModelFreeLinearQuadratic2018,cohenOnlineLinearQuadratic2018,agarwalOnlineControlAdversarial2019,hazanNonstochasticControlProblem2020,yanOnlineNonstochasticControl}, our proposed regret bound no longer explicitly depends on the system dimension due to the use of approximate gradients. Our main contributions are summarized as follows:
    \begin{itemize}
        \item We propose an online control framework that utilizes an approximate model and employs gradient-based online optimization algorithms to learn control policies. This framework allows for significant model uncertainty, which is characterized by the modeling error modulus $\kappa$. We quantitatively analyze the impact of modeling errors on convergence, which is crucial for ensuring the applicability of our algorithms in real-world cyber-physical systems.
        \item We evaluate our conclusions through simulations and in real-world experiments\footnote{See the supplementary video for a demonstration of the experiments: \url{https://youtu.be/OLVvKGba7PA}}. These include a flexible beam, a four-legged walking robot, and a ping-pong playing robot. The ping-pong playing robot is a real-world system that is powered by artificial pneumatic muscles. With our approach, the robotic arm is not only able to continually enhance the tracking performance but also resists the gradually changing dynamic characteristics of the system over time.
        \item We introduce a gradient-based online learning algorithm to learn the control policies, which encompasses both gradient descent and quasi-Newton methods through an appropriate parameter selection. In this algorithm, both the gradient and the quasi-Hessian matrix rely on approximate first-order gradient estimates. The quasi-Newton method demonstrates invariance to linear transformations of the decision variables, thus offering robustness in step size selection and achieving faster convergence compared to the gradient descent method. In contrast, the gradient descent algorithm has a significantly lower computational complexity per iteration, making it more suitable in conjunction with deep neural networks.
        \item We prove convergence guarantees of the algorithms without convexity assumptions and rely only on smoothness assumptions. This ensures that our algorithms and conclusions are indeed applicable to a large class of real-world cyber-physical systems. Our convergence analysis requires an argument for decoupling the approximate Hessian matrix from its dependence on past random variables, which provides an important technical contribution, and achieves a (local) regret bound of $\mathcal{O}\left(T^{\nicefrac{1}{2}}\right)$.
       \item We establish the connection between our algorithms and real-world cyber-physical systems by learning feedforward and feedback controllers in the classic two-degrees-of-freedom control loop in an online manner. We derive the update schemes for our algorithms in both open-loop and closed-loop systems. Furthermore, the effectiveness of both open-loop and closed-loop approaches is evaluated through numerical experiments.
    \end{itemize}
\end{editenv}

\subsection{Structure}
This article follows the structure outlined below: In \secref{sec:problem_formulation}, we will provide a detailed formulation of the stochastic programming problem addressed in this article. Subsequently, we will propose an algorithmic framework to solve this problem and characterize convergence rates in the presence of modeling errors. We will also discuss the assumptions required for proving the convergence results, whereby the detailed proofs are provided in Appendix~\ref{sec:appendixA}. In \secref{sec:interpretation}, we will motivate and discuss the proposed algorithms from the perspective of trust-region methods. \secref{sec:connection} establishes the connection between our algorithms and real-world cyber-physical systems through the classic two-degrees-of-freedom control loop. In addition, we derive the update schemes for our gradient descent algorithm in both open-loop and closed-loop systems. The section highlights the capability of our algorithms to be directly deployed on real-world cyber-physical systems. In \secref{sec:experiments}, our algorithms are applied to various cyber-physical and robotic systems, including a flexible beam, a four-legged walking robot and a ping-pong playing robot. The section further analyzes the tightness of our convergence results and the impact of the modeling errors on convergence. The section also provides detailed information about each example and highlights different methods that can be used for estimating gradients. The article concludes with a summary in \secref{sec:conclusion}.

\section{Problem Setting: Stochastic Online Learning}
\label{sec:problem_formulation}
We consider a specific form of~\eqref{eq:stochastic_setting}, which is tailored to cyber-physical and robotic systems. We incorporate the system dynamics through the mapping $G$ and parameterize actions (or control inputs) $u_t$ via the function $\pi$. Both the dynamics $G$ as well as the action parameterization $\pi$ are added as constraints. Consequently, \eqref{eq:stochastic_setting} is reformulated as follows:
\begin{align}
    \begin{split}
        & \mathbb{E}_{\zeta_t} \left[\sum_{t=1}^T l\left(y_t; \zeta_t\right)\right] -   \min_{\omega \in \Omega} \mathbb{E}_{\zeta_t} \left[\sum_{t=1}^T l\left(y_t^{\star}; \zeta_t\right)\right]\\
        \mathrm{s.t}.~&y_t = G\left(s_0, u_t; \zeta_t\right),~y_t^{\star} = G\left(s_0, u_t^{\star}; \zeta_t\right)\\
        & u_t = \pi\left(\omega_t, y_t; \zeta_t\right),~u_t^{\star} = \pi\left(\omega, y_t^{\star}; \zeta_t\right),~\zeta_t \stackrel{\text{i.i.d.}}{\sim} p_{\zeta},
    \end{split}
    \label{eq:essential_problem}
\end{align}
where the constraints implicitly define $y_t$ as a function of $\omega_t$, $s_0$ and $\zeta_t$, the superscript $\left(\cdot\right)^{\star}$ denotes the optimal value in hindsight. The implicit equation arises due to the fact that feedback loops may potentially be present, and we assume that $y_t$ exists and is well defined. The initial state of the cyber-physical system is denoted by $s_0 \in \mathcal{S}\subset\mathbb{R}^{n_s}$. The vectors $u_t = \left(u_{t,1},\dots,u_{t,q}\right)$ and $y_t=\left(y_{t,1},\dots,y_{t,q}\right)$ denote the input and output sequences of the cyber-physical system, such as a robot, where $u_{t,i}\in\mathcal{U}\subset\mathbb{R}^{m}$ and $y_{t,i}\in\mathcal{Y}\subset\mathbb{R}^{n},~i=1,\dots,q$ represent the input and output at a certain time point $i$ and at iteration $t$ of the learning process. The mapping $G\left(\cdot,\cdot; \zeta\right):\mathcal{S} \times \mathcal{U}^{q} \rightarrow \mathcal{Y}^q$ transforms a sequence of inputs into a sequence of outputs and represents the input-output behavior of the cyber-physical system. The input-output behavior is not necessarily deterministic, due to process, measurement, and actuation uncertainty, which is modeled with the random variable $\zeta_t$. In practice, the mapping $G$ is typically unknown and may exhibit a high degree of nonlinearity, for example due to friction in the joints of a robot. The mapping $\pi\left(\cdot, y_t; \zeta\right): \Omega \rightarrow \mathcal{U}^q$ describes how the decision variables $\omega_t$ affect the controls $u_t$. The feasible set $\Omega$ is assumed to be closed and convex, and the function $l\left(\cdot; \zeta\right):\mathcal{Y}^q \rightarrow \mathbb{R}$ describes the loss function, for example, tracking error, execution time, energy consumption, etc.

\begin{editenv}
    To facilitate understanding for the readers, we provide a detailed explanation of~\eqref{eq:essential_problem} using the drone example mentioned in \secref{sec:motivation}~\citep{hazanIntroductionOnlineControl2025}. In this example, the random variable $\zeta$ represents all uncertain factors, such as terrain, wind, and rainfall, while $G$ denotes the dynamics of the drone in the random environment. As we noted earlier, although indoor environments can be accurately modeled using existing techniques, modeling the dynamics outdoors, considering factors like terrain and weather, is often intractable. The output trajectory $y$ of the drone typically includes its attitude and velocity trajectories, and we do not assume that the state trajectory of the drone is fully observable. The control policy model $\pi$, which we aim to learn online, generates the control policy $u$ based on the drone output trajectory $y$, and may, in general, be affected by the random variable $\zeta$. The loss function $l$ is employed to evaluate the performance of the output trajectory $y$ produced by a given control policy $\pi$; for example, $l$ can be a binary function that determines whether the drone safely reaches its destination, or it can be a nonconvex function derived from a cost map that assesses the quality of the flight path of the drone. Consequently, the regret defined in~\eqref{eq:essential_problem} quantifies the performance difference between the current control policy and the optimal policy in hindsight that would yield the best performance under any environmental conditions. By solving~\eqref{eq:essential_problem}, we obtain a control policy $\pi$ parameterized with the decision variable $\omega$ that enables the drone to perform well across varying terrain and weather conditions.
\end{editenv}

\myedit{Another concrete but simplified example} arises in autonomous driving, where the goal is to control a vehicle to avoid unpredictable hazards, such as sharp turns and obstacles, represented by the random variable $\zeta$. The dynamics of the vehicle, \myedit{unknown to the decision maker}, are denoted by $G$, while the \myedit{control policy} $\pi$ generates control \myedit{actions $u$} for hazard avoidance. The \myedit{partially observable state trajectory} of the vehicle is represented by $y$, including position, speed, direction, etc. The loss function $l$ evaluates the risk faced in avoiding hazards $\zeta$ based on the outputs $y$, with successful avoidance resulting in zero loss. \myedit{The regret defined in~\eqref{eq:essential_problem} characterizes the performance difference between the control policy and the optimal policy that would minimize risk over all unpredictable hazards.} Employing online learning algorithms to solve~\eqref{eq:essential_problem} means that we learn the decision variables $\omega$ so that the controls minimize the risks faced in handling various emergency situations in autonomous driving.

Establishing a precise model of $G$ is often a difficult task. Consequently, in this article, we adopt a black-box representation for $G$, avoiding any explicit characterization of its internal dynamics. Although the dynamic characteristics of $G$ are unknown, we assume that $G$ is differentiable with respect to $u$. Furthermore, we adopt the following notational convention: $\mathcal{G}\left(u_t\right)$ represents an approximation of the gradient of the mapping $G$ at an input location, more precisely, $\mathcal{G}\left(u_t\right)$ denotes an approximation of $\left. \nicefrac{\partial G\left(s_0, u;\zeta\right)}{\partial u} \right|_{u=u_t}$\footnote{The gradient approximation $\mathcal{G}\left(u_t\right)$ may depend on the realization of $\zeta_t$. However, in order to simplify the notation we omit the dependency.}. In the subsequent sections, we will demonstrate that even a rough approximation of $\left. \nicefrac{\partial G\left(s_0, u;\zeta\right)}{\partial u} \right|_{u=u_t}$ can serve as valuable prior knowledge, significantly improving the convergence rate of our algorithms.

To address the optimization problem~\eqref{eq:essential_problem}, we propose~\algoref{algo:online_quasi_newton}, which depending on the choice of $\epsilon$ represents either an online gradient descent or an online quasi-Newton method. We note that $\left(\cdot\right)^{\dagger}$ denotes the Moore-Penrose inverse.
\SetKwInOut{Given}{Given}
\begin{algorithm}[ht]
\SetAlgoLined
\setcounter{AlgoLine}{0}
\KwIn{initial parameters $\omega_1$, constant $\epsilon$ and $\alpha$, iterations $T$, step length $\left\{ \eta_t \right\}^{T}_{t=1}$}
  \For{$t \gets 1$ \KwTo $T$}{
    {\sc Sampling}: $\zeta \sim p_{\zeta} \rightarrow \zeta_t $ \;
    {\sc Implementation}: $ G\left(s_0, \pi \left(\omega_t, y_t; \zeta_t\right); \zeta_t\right)\rightarrow y_t$ \;
    {\sc Evaluation}: $l\left(y_t; \zeta_t\right)$\;
    {\sc Approximation}: $\mathcal{G} \left(u_t\right) \approx \nicefrac{\partial G \left(s_0, \pi\left(\omega_t, y_t;\zeta_t\right);\zeta_t\right)}{\partial u}$\;
    {\sc Calculation}: 
    \begin{flalign*}
    &\mathcal{L}_t =\left(\mathrm{I} - \mathcal{G}\left(u_t\right)\frac{\partial \pi\left(\omega_t,y_t;\zeta_t\right)}{\partial y}\right)^{\dagger} \mathcal{G} \left(u_t\right) \frac{\partial \pi \left(\omega_t,y_t;\zeta_t\right)}{\partial \omega};~\rhd \text{approximation of } \frac{\partial y_t}{\partial \omega_t}&\\
    &\Lambda_t = \frac{1}{\epsilon}\mathcal{L}_{t}^{\text{T}} \nabla^2_y l\left(y_t; \zeta_t\right) \mathcal{L}_{t} + \frac{\alpha}{\epsilon} \nabla_{\omega} \pi\left(\omega_t, y_t;\zeta_t\right) \nabla_{\omega} \pi\left(\omega_t, y_t;\zeta_t\right)^{\text{T}} + \mathrm{I};&\\
    &A_t = \frac{1}{t} \sum_{k=1}^t \Lambda_k;  
    \end{flalign*}
    {\sc Update}: $\omega_{t+1} = \omega_t - \eta_t A_t^{\dagger} \mathcal{L}_t^{\text{T}} \nabla_y l\left(y_t; \zeta_t \right) $ \;
}
\Return{$\omega_{T+1}$}
\caption{{\sc Online-Quasi Newton Method}}
\label{algo:online_quasi_newton}
\end{algorithm}

In principle, the \verb|projection| step from \algoref{algo:online_gradient_descent} could be included to ensure that the parameters $\omega_t$ remain in $\Omega$. However, in cyber-physical applications, $\pi\left(\omega, y; \zeta\right)$ typically represents a neural network, where the parameters have no direct physical interpretation. Constraints often occur on the inputs $u$, and arise, for example, from actuation limits. These can, however, be addressed by parameterizing the neural network in such a way that input constraints are automatically satisfied. In order to simplify the presentation, we will therefore omit the projection onto $\Omega$, or equivalently set $\Omega:= \mathbb{R}^{n_{\omega}}$. We now summarize the convergence guarantees of \algoref{algo:online_quasi_newton} under the following assumptions:

\begin{assumption}[$L$-Smoothness]
\label{asp:smoothness}
Let the loss functions $f\left(\cdot; \zeta\right): \Omega \rightarrow \mathbb{R}$ be $L$-smooth, that is,
\begin{equation*}
    \left|\nabla f \left(v;\zeta\right) - \nabla f \left(\omega;\zeta\right)\right| \leq L \left|v-\omega\right|,
\end{equation*}
for all $\omega, v\in\Omega$.
\end{assumption}

\begin{assumption}[Bounded Variance]
\label{asp:bounded_variance}
There exists a constant $H \geq 0$ such that for all $\omega \in \Omega$ the following inequalities hold:
\begin{equation*}
    \mathbb{E}_{\zeta} \left[\left|\nabla f\left(\omega;\zeta\right) \right|^2 \right] \leq H^2,~\mathbb{E}_{\zeta} \left[\left|\mathcal{F}\left(\omega;\zeta\right) \right|^2 \right] \leq H^2,
\end{equation*}
where $\mathcal{F}\left(\omega;\zeta\right)$ denotes the estimated gradient of $f\left(\omega;\zeta\right)$ induced by $\mathcal{G}\left(u\right)$, while $\nabla f\left(\omega;\zeta\right)$ denotes the true gradient, that is,
\begin{align*}
\begin{split}
    &\mathcal{F} \left(\omega;\zeta\right) = \frac{\partial l\left(y;\zeta\right)}{\partial y} \left(\mathrm{I} - \mathcal{G}\left(u\right)\frac{\partial \pi\left(\omega,y;\zeta\right)}{\partial y}\right)^{\dagger} \mathcal{G} \left(u\right) \frac{\partial \pi \left(\omega,y;\zeta\right)}{\partial \omega},\\
    &\nabla f \left(\omega;\zeta\right) = \frac{\partial l\left(y;\zeta\right)}{\partial y} \left(\mathrm{I} -  \frac{\partial G\left(s_0, u; \zeta\right)}{\partial u}\frac{\partial \pi\left(\omega, y;\zeta\right)}{\partial y} \right)^{\dagger}
    \frac{\partial G\left(s_0, u; \zeta\right)}{\partial u} \frac{\partial \pi\left(\omega, y;\zeta\right)}{\partial \omega}.
\end{split}
\end{align*}
\end{assumption}

\begin{assumption}[Bounded Hessian]\label{asp:bounded_hessian}
Given a sequence of single pseudo-Hessians $\Lambda_t$ obtained according to~\algoref{algo:online_quasi_newton}, there exists a constant $\lambda \geq 1$ such that for all $t=1,\dots,T$ the following inequalities hold:
\begin{equation*}
    1 \leq \lambda_{\mathrm{min}}\left(\Lambda_t\right) \leq \lambda_{\mathrm{max}}\left(\Lambda_t\right) \leq \lambda,
\end{equation*}
where $\lambda_{\mathrm{min}}$ and $\lambda_{\mathrm{max}}$ denote the minimum and maximum eigenvalues of a matrix, respectively.
\end{assumption}

In this work, we abandon the convexity assumption of the objective function $f$ with respect to $\omega$ and employ a more general smoothness assumption instead (see \aspref{asp:smoothness}). \aspref{asp:smoothness} and \aspref{asp:bounded_variance} are standard in non-convex optimization~\citep{OptimizationMethodsLargeScalea}. We note that the non-convexity of the objective function $f$ arises from the nonlinear dynamics of the cyber-physical systems, while the function $l\left(y;\zeta\right)$ can still be chosen to be convex. Thereby, all additive terms in $\Lambda_t$ in \aspref{asp:bounded_hessian} are guaranteed to be positive semi-definite. Furthermore, the matrix $\Lambda_t$ depends on the parameter $\epsilon$, which can always be chosen large enough, such that \aspref{asp:bounded_hessian} is satisfied. Beyond this, we also make the following assumption on the modeling errors of our gradient estimate:
\begin{assumption}[Modeling Error]\label{asp:model_error}
Let the parameters $\omega_t$ evolve according to~\algoref{algo:online_quasi_newton}. There exists a constant $\kappa \in \left[0, 1\right)$ such that for all $t=1,\dots,T$ the following inequality holds:
\begin{equation}
    \left| \mathbb{E}_{\zeta} \left[ \left. \mathcal{F} \left(\omega_t; \zeta\right) \right| \omega_t \right] - \mathbb{E}_{\zeta} \left[ \left. \nabla f \left(\omega_t; \zeta\right) \right|  \omega_t \right] \right|^2 \leq \frac{\kappa^2}{\lambda} \left|\mathbb{E}_{\zeta} \left[ \left. \nabla f \left(\omega_t; \zeta\right) \right| \omega_t \right] \right|^2.
\label{eq:modeling_error}
\end{equation}
\end{assumption}
In fact, the parameter $\lambda$ arises from choosing the $\ell_2$-norm in~\eqref{eq:modeling_error}. If the inequality~\eqref{eq:modeling_error} is expressed in the metric $\left|\cdot\right|_{A_t^{-1}}$, the factor $\nicefrac{1}{\lambda}$ can be avoided, where $|\cdot|_{A^{-1}_t}$ denotes the metric induced by the positive definite matrix $A^{-1}_t$, that is,
\begin{equation*}
 \left| x \right|^2_{A^{-1}_t} := \sup_{\left|x\right| \leq 1} x^{\mathrm{T}} A^{-1}_t x.
\end{equation*}
Then, we have the subsequent conclusions for~\algoref{algo:online_quasi_newton}:
\begin{theorem}\label{thm:online_quasi_newton}
Let the loss functions $f\left(\cdot; \zeta\right):\Omega \rightarrow \mathbb{R}$ satisfy~\aspref{asp:smoothness} and~\aspref{asp:bounded_variance}, and let the pseudo-Hessian $A_t$ satisfy~\aspref{asp:bounded_hessian}. Let the estimate $\mathcal{G}\left(u_t\right)$ satisfy~\aspref{asp:model_error}, and let the step size be chosen as
\begin{equation*}
 \eta =  \sqrt{\frac{ 2 F\left(\omega_{1} \right) }{LH^2T}}.
\end{equation*}
Then the following inequality holds:
\begin{equation}
 \frac{1}{T} \sum_{t=1}^{T}  \mathbb{E}_{\zeta_{1:T}}\left[\left| \nabla F(\omega_t) \right|^2_{A_{t}^{-1}} \right]\leq \sqrt{\frac{2 LH^2 F\left(\omega_{1} \right)}{\left(1-\kappa\right)^2  T}} +\frac{\lambda H^2 \left(\ln{T} + 2\right)}{\left(1 - \kappa \right)  T},
\label{eq:thm1}
\end{equation}
 where $\omega^{\star}:=\argmin_{\omega \in \Omega} F\left(\omega\right)$ denotes the global optimum and $F\left(\omega\right) := \mathbb{E}_{\zeta} \left[f\left(\omega; \zeta\right)\right]$.
\end{theorem}

\begin{figure}[H]
    \centering
    \includegraphics{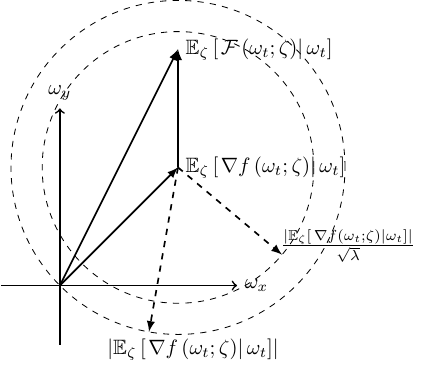}
    \caption{This figure illustrates the geometric meaning of the modeling error modulus $\kappa$ in two-dimensional space. The expectation of the gradient estimate $\mathbb{E}_{\zeta} \left[\left. \mathcal{F}\left(\omega_t;\zeta\right)\right| \omega_t \right]$ lies within the open ball with center $\mathbb{E}_{\zeta} \left[\left. \nabla f\left(\omega_t;\zeta\right)\right| \omega_t \right]$ and radius $\nicefrac{\left|\mathbb{E}_{\zeta} \left[\left. \nabla f\left(\omega_t;\zeta\right)\right| \omega_t \right] \right|}{\sqrt{\lambda}}$.}
    \label{fig:estimate}
\end{figure}

The proof of~\thmref{thm:online_quasi_newton} is included in~Appendix~\ref{sec:appendixA}. From the above conclusion, it is evident that even when using approximate gradients and avoiding convexity assumptions, the expected value of the average of the squared gradients still converges at a rate comparable to many popular stochastic optimization algorithms~\citep{OptimizationMethodsLargeScalea}. We note that, due to the unavailability of $\nicefrac{\partial G\left(s_0, u;\zeta\right)}{\partial u}$ in practical scenarios, the convergence rate of~\algoref{algo:online_quasi_newton} needs to be characterized using the modeling error modulus $\kappa$, and the convergence rate is governed by $\nicefrac{1}{1-\kappa}$. If the modeling error modulus $\kappa$ reaches one, the results become trivial since the right-hand side in~\eqref{eq:thm1} becomes arbitrarily large. When the modeling error modulus $\kappa$ is zero, it implies that the estimate $\mathcal{G}\left(u\right)$ has no bias. The intuitive representation of~\aspref{asp:model_error} in two-dimensional space is illustrated in~\figref{fig:estimate}. The expectation of the gradient estimate $\mathbb{E}_{\zeta} \left[\left. \mathcal{F}\left(\omega_t;\zeta\right)\right| \omega_t \right]$ lies within the open ball with center $\mathbb{E}_{\zeta} \left[\left. \nabla f\left(\omega_t;\zeta\right)\right| \omega_t \right]$ and radius $\nicefrac{\left|\mathbb{E}_{\zeta} \left[\left. \nabla f\left(\omega_t;\zeta\right)\right| \omega_t \right] \right|}{\sqrt{\lambda}}$. This implies that~\aspref{asp:model_error} constrains the estimate $\mathbb{E}_{\zeta} \left[\left. \mathcal{F}\left(\omega_t;\zeta\right)\right| \omega_t \right]$ both in magnitude and direction. Therefore, the parameter $\kappa$ provides a reference for evaluating the quality of the obtained estimates.
   
We observe that by selecting a sufficiently large $\epsilon$, the upper bound $\lambda$ approaches one, thereby transforming~\algoref{algo:online_quasi_newton} from a Newton method to a gradient descent method. This suggests that by adjusting the value of $\epsilon$, we can enable the algorithm to switch between Newton and gradient descent methodologies, leading to the following corollary:
\begin{corollary}\label{cor:guarantee_gradient_descent}
Let the assumptions of~\thmref{thm:online_quasi_newton} be satisfied and let $\epsilon \rightarrow +\infty$. Then, the following inequality holds:
\begin{equation*}
 \frac{1}{T} \sum_{t=1}^{T}  \mathbb{E}_{\zeta_{1:T}}\left[\left| \nabla F(\omega_t) \right|^2 \right]\leq  \frac{\hbar_1}{\sqrt{T}} + \frac{\hbar_2 \ln {T}}{T}  + \frac{2 \hbar_2}{T},
\end{equation*}
where
\begin{equation*}
   \hbar_1 = \frac{\lambda  \sqrt{ 2 LH^2F\left(\omega_{1} \right)}}{ 1-\kappa},~ \hbar_2 = \frac{\lambda^2 H^2 }{1-\kappa}.
\end{equation*}
\end{corollary}
\begin{editenv}
\begin{proof}
    According to \algoref{algo:online_quasi_newton} we know that the matrix $A_t$ is symmetric for all $t \geq 1$. In conjunction with \aspref{asp:bounded_hessian}, we conclude that the eigenvalues of $A_t^{-1}$ are bounded below by $\nicefrac{1}{\lambda}$ and above by unity for all $t \geq 1$. Consequently, the following inequality holds,
    \begin{equation*}
        \left| \nabla F(\omega_t) \right|^2_{A_{t}^{-1}} \geq \frac{1}{\lambda} \left| \nabla F(\omega_t) \right|^2,~\forall t \geq 1.
    \end{equation*}
    We insert the above inequality into~\eqref{eq:thm1}, which yields the following inequality
    \begin{equation*}
        \frac{1}{\lambda} \frac{1}{T} \sum_{t=1}^{T}  \mathbb{E}_{\zeta_{1:T}}\left[\left| \nabla F(\omega_t) \right|^2 \right] \leq \sqrt{\frac{2 LH^2 F\left(\omega_{1} \right)}{\left(1-\kappa\right)^2  T}} +\frac{\lambda H^2 \left(\ln{T} + 2\right)}{\left(1 - \kappa \right)  T},
    \end{equation*}
    and concludes the proof by multiplying by $\lambda$ on both sides.
\end{proof}
\end{editenv}
We note that $A_t$ can be interpreted as an approximation of the Hessian matrix. Using its inverse as the metric to measure the norm of the gradient $\nabla F(\omega_t)$ in~\eqref{eq:thm1}, guarantees the convergence criteria (and the algorithm) are invariant to linear transformations of the parameters $\omega_t$. In the following experiments, we will observe that, compared to the gradient descent method, the quasi-Newton method is better at capturing the local curvature of the objective function, thus achieving more robust convergence and being insensitive to the step size $\eta$.

Next, we will reveal the connection between online learning and stochastic optimization, and provide the corresponding convergence guarantee. Prior to this, we make the following additional assumption:

\begin{editenv}
\begin{assumption}[\L ojasiewicz Inequality]\label{asp:L_inequality}
    There exists a constant $\mu > 0$ and an exponent $\theta \in \left(0, 1\right)$ such that for all $\omega \in \Omega$,
    \begin{equation}
        \left|\nabla F \left(\omega\right)\right|^2 \geq 2 \mu \left(F\left(\omega\right) - F\left(\omega^{\star}\right) \right)^{2\theta}.
    \label{eq:L_inequality}
    \end{equation}
\end{assumption}
Intuitively, this assumption says that near a critical point $\omega^{\star}$, the gradient $\nabla F\left(\omega\right)$ does not become too small relative to how close $F\left(\omega\right)$ is to its value at $\omega^{\star}$. The exponent $\theta$ reflects how ``flat'' or ``steep'' the function is around that point.

Additionally, we give the following definition measuring the distance to the minimum $\omega^{\star}$ in $\Omega$ in terms of function value:
\begin{definition}
    The radius of the minimum $\omega^{\star} \in \Omega$ is defined as
    \begin{equation*}
        \mathcal{D} := \sup_{\widehat{\omega} \in \Omega} \left|F\left(\widehat{\omega}\right) - F\left(\omega^{\star} \right)\right|.
    \end{equation*}
    \label{def:radius}
\end{definition}

Then we show the following lemma:
\begin{lemma}
    Let \aspref{asp:L_inequality} hold, then we have the following conclusions:
    \begin{enumerate}
        \item When $\theta \in \left(0, \frac{1}{2}\right]$, the following inequality holds
        \begin{equation}
             \frac{1}{T} \sum_{t=1}^T \left(F\left(\omega_t\right) - F\left(\omega^{\star}\right)\right) \leq \frac{\mathcal{D}^{1 - 2\theta}}{2 \mu T} \sum_{t=1}^T \left|\nabla F\left(\omega_t\right)\right|^2,
             \label{eq:inequ_1}
        \end{equation}
        where $\mathcal{D}$ is defined in \defref{def:radius}.
        \item When $\theta \in \left(\nicefrac{1}{2}, 1\right)$, the following inequality holds
        \begin{equation}
            \frac{1}{T} \sum_{t=1}^T \left(F\left(\omega_t\right) - F\left(\omega^{\star}\right)\right) \leq \left(\frac{1}{2\mu}\right)^{\frac{1}{2 \theta}}\left(\frac{1}{T} \sum_{t=1}^T \left|\nabla F\left(\omega_t\right)\right|^2\right)^{\frac{1}{2\theta}}.
            \label{eq:inequ_2}
        \end{equation}
    \end{enumerate}
\end{lemma}
\begin{proof}
    For $\theta \in \left(0, \nicefrac{1}{2}\right]$, we note that for all $t\geq 1$ the function $\xi^{2 \theta}$, $\xi \in \left[0, 1\right]$, is concave, which implies that
    \begin{equation*}
        \frac{F\left(\omega_t\right) - F\left(\omega^{\star}\right)}{\mathcal{\mathcal{D}}} \leq \left(\frac{F\left(\omega_t\right) - F\left(\omega^{\star}\right)}{\mathcal{\mathcal{D}}}\right)^{2 \theta},~\forall t\geq 1,
    \end{equation*}
    where we use the fact that 
    \begin{equation*}
        0 \leq \frac{F\left(\omega_t\right) - F\left(\omega^{\star}\right)}{\mathcal{\mathcal{D}}} \leq 1,~\forall t \geq 1.
    \end{equation*}
    Combining with~\eqref{eq:L_inequality} yields
    \begin{equation*}
        \frac{F\left(\omega_t\right) - F\left(\omega^{\star}\right)}{\mathcal{\mathcal{D}}} \leq \frac{1}{\mathcal{D}^{2 \theta}} \left|\nabla F\left(\omega_t\right)\right|^2,~\forall t \geq 1.
    \end{equation*}
    Finally, we conclude the proof of~\eqref{eq:inequ_1} by summing both sides over $t$ from one to $T$.

    For $\theta \in \left(\nicefrac{1}{2}, 1\right)$, we note that for all $t \geq 1$ the function $\xi^{2\theta}$, $\xi \geq 0$, is convex, which yields that
    \begin{equation*}
        \left(\frac{1}{T} \sum_{t=1}^T \left(F\left(\omega_t\right) - F\left(\omega^{\star}\right)\right)\right)^{2 \theta} \leq \frac{1}{T} \sum_{t=1}^T \left(F\left(\omega_t\right) - F\left(\omega^{\star}\right)\right)^{2 \theta}
    \end{equation*}
    by applying Jensen's inequality. Then combining with~\eqref{eq:L_inequality}, we have
    \begin{equation*}
         \left(\frac{1}{T} \sum_{t=1}^T \left(F\left(\omega_t\right) - F\left(\omega^{\star}\right)\right)\right)^{2 \theta} \leq \frac{1}{2 \mu} \frac{1}{T} \sum_{t=1}^T \left|\nabla F\left(\omega_t\right)\right|^2.
    \end{equation*}
    Finally, we conclude~\eqref{eq:inequ_2} by raising both sides to the power of $\nicefrac{1}{2\theta}$.
\end{proof} 

Following \aspref{asp:L_inequality}, we have the conclusions:
\begin{corollary}\label{cor:connection}
   Let the assumptions in~\thmref{thm:online_quasi_newton} be satisfied, and let~\aspref{asp:L_inequality} hold. Then, for any $\epsilon > 0$, the following results hold:
   \begin{enumerate}
       \item When $\theta \in \left(0, \nicefrac{1}{2}\right]$, the expected regret satisfies the inequality:
       \begin{equation}
           \mathbb{E}_{\zeta_{1:T}} \left[ \sum_{t=1}^{T} f \left(\omega_t; \zeta_t\right) \right]- \min_{\omega \in \Omega} 
        \mathbb{E}_{\zeta_{1:T}}  \left[ \sum_{t=1}^{T} f \left(\omega; \zeta_t \right) \right] \leq \frac{\hbar_1 \sqrt{T} + \hbar_2 \ln {T} + 2 \hbar_2}{2\mu  \mathcal{D}^{2\theta - 1}}.
       \label{eq:cor_connection_1}
       \end{equation}
       In addition, let $\delta\in\left(0,1\right)$, then the following inequality on the online regret holds with probability $1-\delta$:
       \begin{equation}
           \sum_{t=1}^{T} f\left(\omega_t; \zeta_t\right) - \min_{\omega \in \Omega} \sum_{t=1}^{T} f \left(\omega; \zeta_t \right) \leq \frac{\hbar_1 \sqrt{T} + \hbar_2 \ln {T} + 2 \hbar_2}{2 \mu \mathcal{D}^{2\theta - 1} \delta}.
       \label{eq:cor_connection_2}
       \end{equation}
       \item When $\theta \in \left(\nicefrac{1}{2}, 1\right)$, the expected regret satisfies the inequality:
       \begin{multline}
            \mathbb{E}_{\zeta_{1:T}} \left[ \sum_{t=1}^{T} f \left(\omega_t; \zeta_t\right) \right]- \min_{\omega \in \Omega} 
            \mathbb{E}_{\zeta_{1:T}}  \left[ \sum_{t=1}^{T} f \left(\omega; \zeta_t \right) \right] \\
            \leq T^{1 - \frac{1}{4\theta}} \left[\left(\frac{\hbar_1}{ 2 \mu}\right)^{\frac{1}{2 \theta}} + 2 \left( \frac{\hbar_2}{2 \mu}\right)^{\frac{1}{2 \theta}} \right],
       \label{eq:cor_connection_21}
        \end{multline}
       In addition, let $\delta\in\left(0,1\right)$, then the following inequality on the online regret holds with probability $1-\delta$:
       \begin{equation}
            \sum_{t=1}^{T} f\left(\omega_t; \zeta_t\right) - \min_{\omega \in \Omega} \sum_{t=1}^{T} f \left(\omega; \zeta_t \right)
           \leq \frac{T^{1 - \frac{1}{4\theta}}}{\delta} \left[\left(\frac{\hbar_1}{ 2 \mu}\right)^{\frac{1}{2 \theta}}  + 2 \left( \frac{\hbar_2}{2 \mu}\right)^{\frac{1}{2 \theta}}\right].
       \label{eq:cor_connection_22}
       \end{equation}
   \end{enumerate}
\end{corollary}

\begin{proof}
    \begin{enumerate}
        \item For the case $\theta \in \left(0, \nicefrac{1}{2}\right]$, we start with~\eqref{eq:inequ_1} and conclude immediately
    \begin{equation*}
        \mathbb{E}_{\omega_{1:T}}\left[ \sum_{t=1}^{T} \mathbb{E}_{\zeta_t} \left[\left. f\left(\omega_t; \zeta_t\right)-f\left(\omega^{\star}; \zeta_t\right)\right|\omega_t \right] \right] \leq \frac{1}{2 \mu \mathcal{D}^{2\theta - 1}} \mathbb{E}_{\omega_{1:T}} \left[ \sum_{t=1}^{T} \left|\nabla F\left(\omega_t\right) \right|^2 \right],
    \end{equation*}
    which leads to the following inequality,
    \begin{equation*}
        \mathbb{E}_{\zeta_{1:T}} \left[ \sum_{t=1}^{T} f\left(\omega_t; \zeta_t\right) -  \sum_{t=1}^{T} f \left(\omega; \zeta_t \right) \right] \leq \frac{\hbar_1 \sqrt{T} + \hbar_2 \ln {T} + 2 \hbar_2}{2 \mu  \mathcal{D}^{2\theta - 1}},~\forall \omega \in \Omega.
    \end{equation*}
    Then, we have
    \begin{equation*}
         \mathbb{E}_{\zeta_{1:T}} \left[ \sum_{t=1}^{T} f \left(\omega_t; \zeta_t\right) \right]- \min_{\omega \in \Omega} 
        \mathbb{E}_{\zeta_{1:T}}  \left[ \sum_{t=1}^{T} f \left(\omega; \zeta_t \right) \right] \leq \frac{\hbar_1 \sqrt{T} + \hbar_2 \ln {T} + 2 \hbar_2}{2\mu  \mathcal{D}^{2\theta - 1}}, 
    \end{equation*}
    which proves~\eqref{eq:cor_connection_1}. Next, we apply Markov's inequality and obtain the following result:
    \begin{equation*}
        \sum_{t=1}^{T} f\left(\omega_t; \zeta_t\right) - \min_{\omega \in \Omega} \sum_{t=1}^{T} f \left(\omega; \zeta_t \right) \leq \frac{\hbar_1 \sqrt{T} + \hbar_2 \ln {T} + 2 \hbar_2}{2 \mu  \mathcal{D}^{2\theta - 1} \delta},
    \end{equation*}
    which holds with probability $1-\delta$. This concludes the proof of the first part of \corref{cor:connection}.
    \item For the case $\theta \in \left(\nicefrac{1}{2}, 1\right)$, we reformulate~\eqref{eq:inequ_2} in the following form
    \begin{equation*}
         \frac{1}{T} \sum_{t=1}^{T} \mathbb{E}_{\zeta_t} \left[\left. f\left(\omega_t; \zeta_t\right)-f\left(\omega^{\star}; \zeta_t\right)\right|\omega_t \right] \leq \left(\frac{1}{2\mu}\right)^{\frac{1}{2 \theta}}\left(\frac{1}{T} \sum_{t=1}^T \left|\nabla F\left(\omega_t\right)\right|^2\right)^{\frac{1}{2\theta}}.
    \end{equation*}
    Then we take expectation on both sides to obtain the following inequality
    \begin{multline*}
        \frac{1}{T} \mathbb{E}_{\omega_{1:T}} \left[ \sum_{t=1}^{T} \mathbb{E}_{\zeta_t} \left[\left. f\left(\omega_t; \zeta_t\right)-f\left(\omega^{\star}; \zeta_t\right)\right|\omega_t \right] \right] \\ 
        \leq \left(\frac{1}{2\mu}\right)^{\frac{1}{2 \theta}} \mathbb{E}_{\omega_{1:T}} \left[ \left(\frac{1}{T} \sum_{t=1}^T \left|\nabla F\left(\omega_t\right)\right|^2\right)^{\frac{1}{2\theta}} \right].
    \end{multline*}
    We apply Jensen's inequality for concave functions to the right-hand side of the above inequality, and conclude
    \begin{multline*}
        \mathbb{E}_{\omega_{1:T}} \left[ \left(\frac{1}{T} \sum_{t=1}^T \left|\nabla F\left(\omega_t\right)\right|^2\right)^{\frac{1}{2\theta}} \right] \leq \left(\frac{1}{T}  \mathbb{E}_{\omega_{1:T}} \left[\sum_{t=1}^T \left|\nabla F\left(\omega_t\right)\right|^2 \right] \right)^{\frac{1}{2\theta}} \\
        \leq \left( \frac{\hbar_1 }{ \sqrt{T}} + \frac{\hbar_2 \ln{T}}{T} + \frac{2 \hbar_2}{T} \right)^{\frac{1}{2 \theta}} \\
        \leq \left(\frac{\hbar_1 }{ \sqrt{T}}\right)^{\frac{1}{2 \theta}} + \left( \frac{\hbar_2 \left(\ln{T} + 2\right) }{T}\right)^{\frac{1}{2 \theta}},
    \end{multline*}
    where the last inequality is guaranteed by the subadditivity of concave functions. 
    
    Next, we take the leftmost and rightmost sides of the above inequalities, and for all $\omega \in \Omega$ have
    \begin{multline*}
         \mathbb{E}_{\zeta_{1:T}} \left[ \sum_{t=1}^{T} f\left(\omega_t; \zeta_t\right) -  \sum_{t=1}^{T} f \left(\omega; \zeta_t \right) \right] \\
         \leq T^{1 - \frac{1}{4\theta}} \left[\left(\frac{\hbar_1}{ 2 \mu}\right)^{\frac{1}{2 \theta}}  + \left( \frac{\hbar_2}{2 \mu}\right)^{\frac{1}{2 \theta}} \left(\ln{T} + 2\right)^{\frac{1}{2 \theta}} T^{- \frac{1}{4 \theta}}\right] \\ 
         \leq  T^{1 - \frac{1}{4\theta}} \left[\left(\frac{\hbar_1}{ 2 \mu}\right)^{\frac{1}{2 \theta}}  + 2 \left( \frac{\hbar_2}{2 \mu}\right)^{\frac{1}{2 \theta}} \right],
    \end{multline*}
    where the last inequality is based on the fact that for all $\theta \in \left(\nicefrac{1}{2}, 1\right)$ and $T \geq 1$ we have
    \begin{equation*}
         \left(\ln{T} + 2\right)^{\frac{1}{2 \theta}} T^{- \frac{1}{4 \theta}} \leq 2.
    \end{equation*}
    Furthermore, we have
    \begin{multline*}
        \mathbb{E}_{\zeta_{1:T}} \left[ \sum_{t=1}^{T} f \left(\omega_t; \zeta_t\right) \right]- \min_{\omega \in \Omega} 
        \mathbb{E}_{\zeta_{1:T}}  \left[ \sum_{t=1}^{T} f \left(\omega; \zeta_t \right) \right] \\
        \leq T^{1 - \frac{1}{4\theta}} \left[\left(\frac{\hbar_1}{ 2 \mu}\right)^{\frac{1}{2 \theta}}  + 2 \left( \frac{\hbar_2}{2 \mu}\right)^{\frac{1}{2 \theta}} \right],
    \end{multline*}
    which proves~\eqref{eq:cor_connection_21}. By applying Markov's inequality we obtain the following result:
    \begin{equation*}
        \sum_{t=1}^{T} f\left(\omega_t; \zeta_t\right) - \min_{\omega \in \Omega} \sum_{t=1}^{T} f \left(\omega; \zeta_t \right)
       \leq \frac{T^{1 - \frac{1}{4\theta}}}{\delta} \left[\left(\frac{\hbar_1}{ 2 \mu}\right)^{\frac{1}{2 \theta}}  + 2 \left( \frac{\hbar_2}{2 \mu}\right)^{\frac{1}{2 \theta}} \right],
    \end{equation*}
    which holds with probability $1 - \delta$. This concludes the proof of the second part of \corref{cor:connection}.
    \end{enumerate}    
\end{proof}

The results of \corref{cor:connection} highlight that when $\theta \in \left(0, \nicefrac{1}{2}\right]$, \algoref{algo:online_quasi_newton} is expected to converge at the standard stochastic rate of $\mathcal{O}\left(\nicefrac{1}{\sqrt{T}}\right)$ (or equivalently $\mathcal{O}(\sqrt{T})$ regret). In contrast, the regret is upper bounded by $\mathcal{O}\left(T^{1-\nicefrac{1}{4\theta}}\right)$ when $\theta \in \left(\nicefrac{1}{2}, 1\right)$. In both cases, the convergence rates are dimension-independent, and the errors in the gradient estimate contribute a factor of $\nicefrac{1}{1-\kappa}$.
\end{editenv}

\section{Interpretation of \algoref{algo:online_quasi_newton} as Trust-Region Approach}
\label{sec:interpretation}
In this section, we will interpret~\algoref{algo:online_quasi_newton} from the perspective of a trust-region method. However, we only demonstrate the key steps, while the more detailed derivation is included in Appendix~\ref{sec:app_trust_region}. For a fixed sample $\zeta$, the loss function $f \left(\omega;\zeta \right)$ and the parameterized network $\pi\left(\omega, y; \zeta\right)$ will be denoted as $f \left(\omega\right)$ and $\pi\left(\omega\right)$ for notational convenience. A trust-region method solves the following transformed optimization problem at each step~\citep{yuanRecentAdvancesTrust2015, martinezLocalMinimizersQuadratic1994, sorensenNewtonMethodModel1982}:
\begin{align*}
   \min_{d \in \Omega}~& f \left(\omega \right) + \nabla f \left(\omega\right)^{\text{T}}  d +\frac{1}{2} d^{\text{T}} \nabla^2 f\left(\omega\right) d\\
   \text{s.t.}~&\left|d\right| \leq \Delta,
\end{align*}
where $\Delta$ denotes the radius of the trust region. Drawing inspiration from this idea, we approximate the loss function locally as follows:
\begin{multline}
M\left(v, \omega \right) = f \left(\omega\right) + \left(v-\omega \right)^{\text{T}} \mathcal{L}^{\text{T}} \nabla_y l\left(y;\zeta\right)  +\frac{1}{2} \left(v-\omega\right)^{\text{T}} \mathcal{L}^{\text{T}} \nabla^2_y l\left(y;\zeta\right) \mathcal{L} \left(v-\omega\right)\\
+\underbrace{\frac{\alpha}{2} \left|\nabla_{\omega} \pi \left(\omega; \zeta\right)^{\text{T}} \left(v-\omega \right) \right|^2}_{\text{Term 1}} + \underbrace{\frac{\epsilon}{2} \left|v-\omega\right|^2}_{\text{Term 2}},
\label{eq:intermediate_trust_region}
\end{multline}
where
\begin{equation*}
    \mathcal{L}=\frac{\partial y}{\partial \omega} = \left(\mathrm{I} - \mathcal{G}\left(u\right)\frac{\partial \pi\left(\omega,y;\zeta\right)}{\partial y}\right)^{\dagger} \mathcal{G} \left(u\right) \frac{\partial \pi \left(\omega,y;\zeta\right)}{\partial \omega}.
\end{equation*}
We note that Term $1$ and Term $2$ are used as penalty terms to ensure that when $v \in \Omega$ deviates from the parameters $\omega$, the value of $M\left(v, \omega\right)$ remains strictly greater than that of $f\left(\omega\right)$. Furthermore, these terms help prevent drastic changes in the parameters $\omega$ and their corresponding outputs $u=\pi \left(\omega\right)$ during each iteration of the resulting iterative scheme. In our setting, we can ensure that the trust region constraint is always satisfied, specifically $\left|v - \omega\right| \leq \Delta$, by appropriately adjusting the constant $\epsilon$ in~\eqref{eq:intermediate_trust_region}.

Next, we calculate the closed-form solution of the minimum point of $\mathbb{E}_{\zeta} \left[\left. M\left(v, \omega\right)\right| \omega \right]$ as follows:
\begin{equation}
\omega^{\star} = \omega - \frac{1}{\epsilon} \left(\Sigma_{1} + \alpha \Sigma_{2} + \text{I} \right)^{-1}  \int \mathcal{L} \left(\xi\right)^{\text{T}} \nabla_y l\left(y;\xi\right) p_{\zeta} \left(\xi\right) \text{d} \xi,
\label{eq:closed_form_solution}
\end{equation}
where
\begin{multline*}
    \Sigma_{1} = \int  \mathcal{L} \left(\xi\right)^{\text{T}} \nabla^2_y l\left(y; \xi\right) \mathcal{L} \left(\xi\right) p_{\zeta}\left(\xi\right) \text{d} \xi,~\Sigma_{2} = \int \nabla_{\omega} \pi\left(\omega;\xi\right) \nabla_{\omega} \pi \left(\omega;\xi\right)^{\text{T}}~p_{\zeta} \left(\xi\right) \text{d} \xi,
\end{multline*}
and $\text{I} \in \mathbb{R}^{n_{\omega} \times n_{\omega}}$ denotes the identity matrix. Since the trust-region function $M\left(v, \omega\right)$ provides only a local approximation of the cost function $f\left(\omega\right)$ in the vicinity of the point $\omega$, we adopt the following iterative scheme to guide the parameters $\omega$ towards the optimal value:
\begin{equation*}
\omega_{t+1} = \omega_{t} -\frac{1}{\epsilon} \left(\Sigma_{1} + \alpha \Sigma_{2} + \text{I} \right)^{-1}  \int \mathcal{L} \left(\xi\right)^{\text{T}} \nabla_y l\left(y_t;\xi\right) p_{\zeta} \left(\xi\right) \text{d} \xi,~t=1,\dots,T.
\end{equation*}
In the context of stochastic optimization, where global information is lacking, the matrices $\Sigma_1$ and $\Sigma_2$ are unavailable. Therefore, we employ the following estimates to respectively replace $\Sigma_1$ and $\Sigma_2$ for $t=1,\dots,T$:
\begin{align*}
\Sigma_{1}^{t} = \frac{1}{\epsilon t} \sum_{k=1}^{t}\mathcal{L}_{k}^{\text{T}} \nabla^2_y l\left(y_k;\zeta_k\right) \mathcal{L}_{k},~\Sigma_{2}^{t} = \frac{1}{\epsilon t} \sum_{k=1}^{t} \nabla_{\omega} \pi \left(\omega_k, y_k;\zeta_k\right)  \nabla_{\omega} \pi \left(\omega_k, y_k;\zeta_k\right)^{\text{T}},
\end{align*}
Likewise, we replace the term $
\int \mathcal{L} \left(\xi\right)^{\text{T}} \nabla_y l\left(y_t; \xi\right) p_{\zeta} \left(\xi\right) \text{d} \xi$ with the estimate $\mathcal{L}_{t}^{\text{T}} \nabla_y l\left(y_t; \zeta_t\right)$. Finally, we introduce the step size $\eta_t$ at each iteration and get the following iterative scheme:
\begin{equation}
\omega_{t+1} = \omega_t - \eta_t \left(\Sigma_{1}^t+\alpha \Sigma_{2}^t + \text{I} \right)^{-1} \mathcal{L}_{t}^{\text{T}} \nabla_y l\left(y_t;\zeta_t\right).
\label{eq:final_iterative_scheme}
\end{equation}

We note that in~\eqref{eq:final_iterative_scheme}, the term $\nabla_y l\left(y_t; \zeta_t\right)$ can be easily obtained at each iteration by observing the output of the cyber-physical system. Meanwhile, $\Sigma^t_2$ can be obtained by performing backpropagation at each iteration. Therefore, the only term that is difficult to obtain accurately in \algoref{algo:online_quasi_newton} is the gradient $\mathcal{L}_t$ and the corresponding $\Sigma^t_1$. Therefore, we use the approximate gradient $\mathcal{G}\left(u\right)$ instead of $\nicefrac{\partial G\left(s_0, u;\zeta\right)}{\partial u}$.

\section{Connection to Cyber-Physical Systems and Robotics}
\label{sec:connection}
In this article, we not only emphasize the importance of convergence analysis but also bridge the gap between theory and practical cyber-physical systems. From the analysis in~\secref{sec:problem_formulation} we conclude that as long as the approximate gradient $\mathcal{G}\left(u_t\right)$ is accurate enough (as characterized by the error modulus $\kappa$) convergence is guaranteed and~\algoref{algo:online_quasi_newton} can be directly deployed on a cyber-physical system. We note that for cyber-physical systems, common estimation methods include system identification and finite difference estimation~\citep{ljungPerspectivesSystemIdentification2010, pintelonSystemIdentificationFrequency2012a, careFiniteSampleSystemIdentification2018, tsiamisFiniteSampleAnalysis2019, campiFiniteSampleProperties2002}. In the following examples, we will use these estimation methods to demonstrate the broad applicability and robustness (insensitivity to modeling errors) of our algorithm.

First of all, we endow the random variable $\zeta$ with a specific meaning for cyber-physical systems. In the following sections of this article, we consider the random variable $\zeta$ as the reference trajectory $y_{\mathrm{ref}} \in \mathcal{Y}^q$ that the system is required to track. Given that the learning task is trajectory tracking, an intuitive choice for the loss function $l$ is the deviation from the reference trajectory, that is,
\begin{equation}
l\left(y_t; y_{\text{ref},t} \right) := \frac{1}{2} \left| G \left(s_0, u_t; y_{\text{ref}, t} \right) - y_{\text{ref}, t} \right|^2,~y_{\text{ref}, t} \stackrel{\text{i.i.d.}}{\sim} p_{y_{\text{ref}}},~t=1,\dots,T.
\label{eq:tracking_error}
\end{equation}
Thereby, we have $\nabla^2_y l\left(y;\zeta\right) = \mathrm{I}$. As delineated in~\eqref{eq:stochastic_setting}, $t$ denotes the current iteration, with the total iterations denoted by $T$. This notation inherently suggests that the reference trajectory $y_{\text{ref}}$ evolves in correspondence with the progression of iterations. Concurrently, at each iteration, the reference trajectory is randomly sampled from a fixed yet unknown distribution $p_{y_{\text{ref}}}$. The distribution $p_{y_{\text{ref}}}$ depends upon the specific robotic task at hand, with trajectory tracking serving as a broad, universal objective. For example, in the context of training a robot for table tennis, $p_{y_{\text{ref}}}$ is determined by the trajectories experienced by the end-effector during ball interception~\citep{maLearningbasedIterativeControl2022b, ReinforcementLearningModelbased, tobuschatDataEfficientOnlineLearning2023}. Given the specific meaning of $\zeta$ and the action parameterization $\pi$, implies that we have identified a nonlinear feedforward and feedback controller that yields accurate trajectory tracking for any $y_{\text{ref}} \sim p_{y_{\text{ref}}}$ when solving~\eqref{eq:essential_problem}. For simplification and without really compromising generality (we could extend the function $\pi$ to also account for $s_0$), we assume that the system consistently initializes from an identical state prior to each iteration. As such, the initial state $s_0$ can be fixed and omitted. For example, in the experiments with the ping-pong robot we drive the robot back to a rest position after each iteration of the online learning with a simple proportional-integral-derivative (PID) controller.

\begin{figure}[ht]
\centering
\begin{subfigure}{\textwidth}
    \centering
    \includegraphics{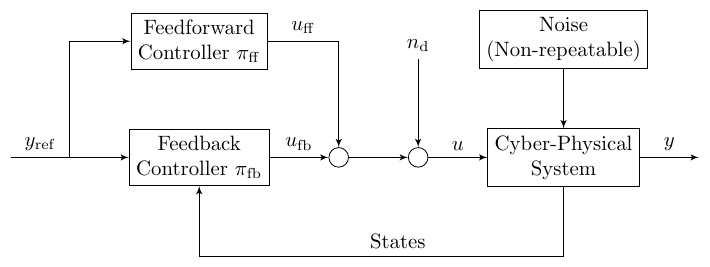}
    \caption{two-degrees-of-freedom control loop}
    \label{fig:two_degree}
\end{subfigure}
\begin{subfigure}{\textwidth}
    \centering
    \includegraphics{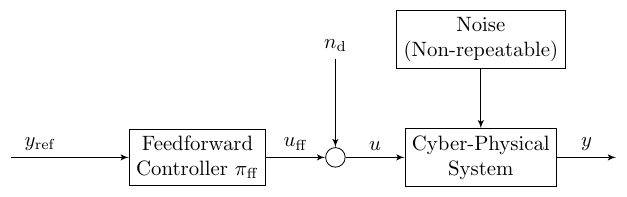}
    \caption{feedforward control loop}
    \label{fig:one_degree}
\end{subfigure}
\caption{The figure shows the classical two-degrees-of-freedom control framework in panel (a), which includes a feedforward controller and a feedback controller, and a pure feedforward control framework in panel (b). The variable $n_{\text{d}}$ denotes a disturbance, which will subsequently be used to obtain an approximate gradient $\mathcal{G}\left(u_t\right)$.}
\label{fig:control_loop}
\end{figure}

To address the learning tasks in cyber-physical systems, we draw inspiration from the classical two-degrees-of-freedom control framework, which includes a feedforward block and a feedback block, and is shown in~\figref{fig:two_degree}. The two-degrees-of-freedom control loop has extensive applications in machine learning within the context of robotics. We can contrast our approach of learning feedforward and feedback controllers to RL, where the objective is to learn a feedback controller (policy) that minimizes a designated reward function~\citep{suttonReinforcementLearningSecond2018a, liDeepReinforcementLearning2018a}. In contrast to RL, which is often based on approximately solving the Bellman equation, we do not use any dynamic programming strategy in our approach.

In our approach, we use $\pi_{\text{ff}}$ and $\pi_{\text{fb}}$ to represent the parameterized feedforward and feedback networks, respectively, and $\omega_{\text{ff}}$ and $\omega_{\text{fb}}$ to denote their corresponding parameters. Then, the relation between $y_t$ and $u_t$ in~\eqref{eq:essential_problem} can be reformulated as follows:
\begin{align}
    \begin{split}
        & y_t = G\left(s_0, u_t;  \zeta_{t}\right),\\
        & u_t = \pi_{\text{ff}} \left(\omega_{\text{ff}, t}; \zeta_{t}\right) + \pi_{\text{fb}} \left(\omega_{\text{fb}, t};y_t - \zeta_{t} \right),~\zeta_{t} \stackrel{\text{i.i.d.}}{\sim} p_{\zeta},
    \end{split}
    \label{eq:piff_pifb}
\end{align}
that is, the input $u_t$ is the combination of a feedforward part $\pi_{\text{ff}}$ that does not depend on $y_t$ and a feedback part $\pi_{\text{fb}}$ that depends on the deviation of $y_t$ from the reference trajectory $\zeta_t$. Due to the inclusion of feedback $\pi_{\text{fb}}$, the calculation of the gradient in~\algoref{algo:online_quasi_newton} becomes more complex and less intuitive compared to the open-loop situation where $\pi_{\text{fb}} = 0$. Hence, we will demonstrate the computation of the gradients $\nicefrac{\partial y}{\partial \omega_{\text{ff}}}$ and $\nicefrac{\partial y}{\partial \omega_{\text{fb}}}$ in the closed-loop system and discuss their implications. The critical aspect to note at this point is that~\eqref{eq:piff_pifb} defines an implicit equation for $y_t$ and also $u_t$. We should therefore think of $y_t$ and $u_t$ as functions of $\omega_{\text{ff}}$, $\omega_{\text{fb}}$ and $\zeta_t$, that is, $u = u\left(\omega_{\text{ff}},\omega_{\text{fb}}; \zeta\right)$ and $y = y\left(\omega_{\text{ff}},\omega_{\text{fb}}; \zeta\right)$. The gradient of the loss function $f$ with respect to the parameters can be calculated as follows:
\begin{equation*}
    \nabla_{\omega} f \left(\omega; \zeta\right) = 
    \begin{bmatrix}
        \frac{\partial y^{\text{T}}}{\partial \omega_{\text{ff}}} &\frac{\partial y^{\text{T}}}{\partial \omega_{\text{fb}}}
    \end{bmatrix}
    \nabla_{y} l\left(y;\zeta\right).
\end{equation*}
By combining the two equations in~\eqref{eq:piff_pifb} we get a single implicit equation for $y$. The differential $\nicefrac{\partial y}{\partial \omega_{\text{ff}}}$ can now be obtained by differentiating the implicit equation with respect to $\omega_{\text{ff}}$ (implicit function theorem):
\begin{multline*}
    \frac{\partial y}{\partial \omega_{\text{ff}}} =  \left. \frac{\partial G\left(u; \zeta\right) }{\partial u}\right|_{u=\pi_{\text{ff}} + \pi_{\text{fb}}} \left. \frac{\partial \pi_{\text{fb}}\left(\omega_{\text{fb}}; \zeta^{\circ} \right)  }{\partial \zeta^{\circ}}\right|_{\zeta^{\circ} = G\left(u; \zeta\right)-\zeta} \frac{\partial y}{\partial \omega_{\text{ff}}} \\
    + \left. \frac{\partial G\left(u; \zeta\right) }{\partial u}\right|_{u=\pi_{\text{ff}} + \pi_{\text{fb}}}  \frac{\partial \pi_{\text{ff}}\left(\omega_{\text{ff}};\zeta\right) }{\partial \omega_{\text{ff}}}.
\end{multline*}
This can be rearranged to
\begin{multline}
\frac{\partial y}{\partial \omega_{\text{ff}}} = \left(\mathrm{I} - \frac{\partial G\left(u;\zeta\right)}{\partial u} \left. \frac{\partial \pi_{\text{fb}}\left(\omega_{\text{fb}};\zeta^{\circ}\right)}{\partial \zeta^{\circ}}\right|_{\zeta^{\circ} = G\left(u; \zeta\right)-\zeta} \right)^{\dagger}\\
\frac{\partial G\left(u;\zeta\right)}{\partial u} \frac{\partial \pi_{\text{ff}}\left(\omega_{\text{ff}};\zeta\right)}{\partial \omega_{\text{ff}}}.
\label{eq:closed_loop1}
\end{multline}
The expression $\nicefrac{\partial y}{\partial \omega_{\text{fb}}}$ can be derived with a similar argument and results in
\begin{multline}
 \frac{\partial y}{\partial \omega_{\text{fb}}} =  \left(\mathrm{I} - \frac{\partial G\left(u;\zeta\right)}{\partial u} \left. \frac{\partial \pi_{\text{fb}}\left(\omega_{\text{fb}};\zeta^{\circ}\right)}{\partial \zeta^{\circ}}\right|_{\zeta^{\circ} = G\left(u; \zeta\right)-\zeta} \right)^{\dagger}\\
 \frac{\partial G\left(u;\zeta\right)}{\partial u} \left. \frac{\partial \pi_{\text{fb}}\left(\omega_{\text{fb}};\zeta^{\circ}\right)}{\partial \omega_{\text{fb}}}\right|_{\zeta^{\circ} = G\left(u; \zeta\right)-\zeta}.
 \label{eq:closed_loop2}
\end{multline}

We observe that the term $\nicefrac{\partial G\left(u;\zeta\right)}{\partial u}$ consistently represents the gradient of the open-loop system and, as previously mentioned, can be approximated using the estimate $\mathcal{G}\left(u\right)$. This approximation renders the terms $\nicefrac{\partial y}{\partial \omega_{\text{ff}}}$ and $\nicefrac{\partial y}{\partial \omega_{\text{fb}}}$ computable. We also note that the feedback controller may reduce the effect of estimation errors in $\mathcal{G}$ on the resulting gradient estimates $\mathcal{F}$. Indeed, if $\nicefrac{\partial \pi_{\text{fb}} \left(\omega_{\text{fb}};\zeta^{\circ} \right)}{\partial \zeta^{\circ}}$ is large, both expressions reduce to 
\begin{equation*}
    \frac{\partial \pi_{\text{fb}} \left(\omega_{\text{fb}};\zeta^{\circ} \right)^{\dagger}}{\partial \zeta^{\circ}}
    \frac{\partial \pi_{\text{fb}} \left(\omega_{\text{fb}};\zeta^{\circ} \right)}{\partial \omega_{\text{fb}}},~\frac{\partial \pi_{\text{fb}} \left(\omega_{\text{fb}};\zeta^{\circ} \right)^{\dagger}}{\partial \zeta^{\circ}}
    \frac{\partial \pi_{\text{ff}} \left(\omega_{\text{ff}};\zeta\right)}{\partial \omega_{\text{ff}}},
\end{equation*}
respectively, which means that $\nabla f$ is approximately independent of $G$ for large $\nicefrac{\partial \pi_{\text{fb}} \left(\omega_{\text{fb}};\zeta^{\circ} \right)}{\partial \zeta^{\circ}}$. If the feedback gain is small, however, $\nicefrac{\partial y}{\partial \omega_{\text{ff}}}$ and $\nicefrac{\partial y}{\partial \omega_{\text{fb}}}$ reduce to 
\begin{equation*}
    \frac{\partial G\left(s_0, u;\zeta\right)}{\partial u} \frac{\pi_{\text{ff}} \left( \omega_{\text{ff}};\zeta \right) }{\partial \omega_{\text{ff}}},~\frac{\partial G\left(s_0, u;\zeta\right)}{\partial u} \frac{\pi_{\text{fb}}\left( \omega_{\text{fb}};\zeta^{ \circ} \right) }{\partial \omega_{\text{fb}}}.
\end{equation*}
Moving forward, we will briefly show that the term
\begin{equation}
     \left(\mathrm{I} - \frac{\partial G\left(u;\zeta\right)}{\partial u} \frac{\partial \pi_{\text{fb}}\left(\omega_{\text{fb}};\zeta^{\circ}\right)}{\partial \zeta^{\circ}} \right)^{\dagger} \frac{\partial G\left(u;\zeta\right)}{\partial u}
\label{eq:closed_loop3}
\end{equation}
is the gradient of the closed-loop system with respect to the external input $n_{\text{d}}$ (see~\figref{fig:control_loop}). This finding enables us to directly derive gradient estimation approaches for the closed-loop system, which will be denoted as $\mathcal{G}^{\circ} \left(\omega_{\text{ff}}, \omega_{\text{fb}},\zeta\right)$. We perform the following calculations (implicit function theorem):
\begin{multline*}
     \frac{\partial y}{\partial n_{\text{d}}} = \left. \frac{\partial G\left(u;\zeta\right)}{\partial u} \right|_{u=\pi_{\text{ff}}+\pi_{\text{fb}}} \left. \frac{\partial \pi_{\text{fb}}\left(\omega_{\text{fb}}; \zeta^{\circ}\right) }{\partial \zeta^{\circ}} \right|_{\zeta^{\circ} = G\left(u; \zeta\right)-\zeta}  \frac{\partial y}{\partial n_{\text{d}}}\\
     +  \left. \frac{\partial G\left(u;\zeta\right)}{\partial u} \right|_{u=\pi_{\text{ff}}+\pi_{\text{fb}}},
\end{multline*}
which results in
\begin{equation*}
    \frac{\partial y}{\partial n_{\text{d}}} =  \left(\mathrm{I} - \frac{\partial G\left(u;\zeta\right)}{\partial u} \frac{\partial \pi_{\text{fb}}\left(\omega_{\text{fb}};\zeta^{\circ}\right)}{\partial \zeta^{\circ}} \right)^{\dagger} \frac{\partial G\left(u;\zeta\right)}{\partial u}.
\end{equation*}
Intuitively, $\nicefrac{\partial y}{\partial n_{\text{d}}}$ describes the sensitivity of $y$ in closed-loop to changes in $u$.

Consequently, the terms $\nicefrac{\partial y}{\partial \omega_{\text{ff}}}$ and $\nicefrac{\partial y}{\partial \omega_{\text{fb}}}$, apart from being derived from~\eqref{eq:closed_loop1} and~\eqref{eq:closed_loop2}, can also be obtained through the following more direct approach
\begin{equation*}
     \frac{\partial y}{\partial \omega_{\text{ff}}} = \mathcal{G}^{\circ} \left(\omega_{\text{ff}},\omega_{\text{fb}},\zeta\right) \frac{\partial \pi_{\text{ff}}\left(\omega_{\text{ff}};\zeta\right) }{\partial \omega_{\text{ff}}},~\frac{\partial y}{\partial \omega_{\text{fb}}} = \mathcal{G}^{\circ} \left(\omega_{\text{ff}},\omega_{\text{fb}},\zeta\right) \frac{\partial \pi_{\text{fb}}\left(\omega_{\text{fb}};\zeta^{\circ}\right) }{\partial \omega_{\text{fb}}},
\end{equation*}
thereby allowing for direct computations if $\mathcal{G}^{\circ}$ is known. As we will highlight with simulation experiments $\mathcal{G}^{\circ}\left(\omega_{\text{ff}},\omega_{\text{fb}},\zeta\right)$ can be computed directly by performing stochastic rollouts with different random perturbations $n_{\text{d}}$ (see \secref{sec:ant}). 

\section{Experiments}
\label{sec:experiments}
In this section, we will demonstrate the effectiveness of our algorithms through extensive experiments conducted on various cyber-physical systems, including simulation and real-world experiments. Additionally, we evaluate the accuracy of the theoretical convergence rate of the algorithm proposed in~\secref{sec:problem_formulation}. In each experiment, we will first introduce the underlying cyber-physical system. Then, based on the specific systems, we will describe the distribution of the reference trajectories, the structure of the parameterized models $\pi$, and the corresponding structure of the inputs. We highlight that even shallow networks work well with our algorithms. Depending on different scenarios, we will employ an appropriate method to obtain gradient estimates $\mathcal{G}\left(u\right)$ and $\mathcal{G}^{\circ}$. Finally, we will discuss the results of the experiments in terms of convergence rate and robustness to modeling errors. It is worth mentioning that the primary intention of the following experiments is not to outperform any existing (reinforcement) learning algorithms but to evaluate the effectiveness of our algorithms and offer another possibility for the development of machine learning in the field of robotics.

\subsection{Cantilever Beam}
\label{sec:numerical_experiment}
\begin{figure}[H]
\centering
\includegraphics{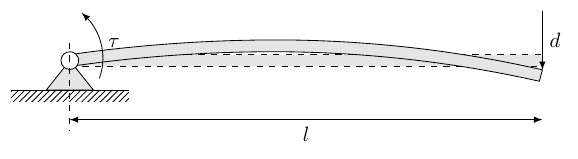}
\caption{Deformation of the cantilever beam under the active torque and an external disturbance $d$, where the dashed line represents the position of the cantilever beam when at rest.}
\label{fig:beam}
\end{figure}

The first example is based on the control of a flexible cantilever beam. The example illustrates how our approach can easily handle dynamical systems with a large number of hidden states (here $\num{100}$). We note that current RL algorithms have difficulties in dealing with continuous state and action spaces exceeding two dozen states. Modeling and controlling flexible structures has numerous engineering applications~\citep{shabanaFlexibleMultibodyDynamics1997, amiroucheFundamentalsMultibodyDynamics2007}, such as active vibration control of wind turbines, aircraft wings or turbo generator shafts. We consider a cantilever beam illustrated in \figref{fig:beam}, where the left end of the beam is hinged to a joint, and the active torque $\tau$ is applied only at the left end to counteract the disturbances at the tip of the flexible body. The total length of the entire cantilever beam in a rest configuration is denoted by $l$.

\begin{figure}[ht]
\centering
\includegraphics{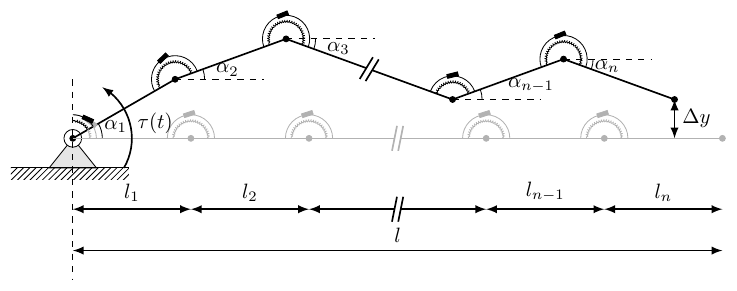}
\caption{The figure depicts the discrete model of the cantilever beam obtained using the lumped-parameter method. The entire beam is decomposed into $n$ rigid units, with adjacent units interconnected by a pair of spring and damper. Each rigid unit has a length of $l_i$, and the angle it makes with the horizontal plane is denoted by $\alpha_i$. The active torque is applied only to the first rigid unit hinged to the wall. The gray object indicates the position of the beam when at rest.}
\label{fig:dis_beam}
\end{figure}

We employ a lumped-parameter method to discretize the cantilever beam, as illustrated in \figref{fig:dis_beam}, into a collection of $n$ rigid units. 
In the discrete model of the cantilever beam, the entire beam is divided into $n$ rigid units, either uniformly or non-uniformly. These rigid units are coupled through joints equipped with internal springs and dampers. The joints provide the degrees of freedom required for deformation. The mass, spring, and damper elements offer inertial, restorative, and dissipative forces, respectively, which collectively account for the deformation. Typically, due to the small deformation assumption, the springs and dampers inside each joint are considered linear. In this experiment, however, we will explore the performance of our algorithms in nonlinear systems. Therefore, we adjust the parameters of the cantilever beam (such as the moment of inertia and damping parameters) to enable large deformations and introduce the following nonlinear spring force:
\begin{equation*}
    f_{\mathrm{spring}, i}\left(\Delta \alpha_i\right) = k_{i1} \Delta \alpha_i + k_{i2} \Delta \alpha_i^3 + k_{i3} \Delta \alpha_i^5,~i=1,\dots,n,
\end{equation*}
where $\Delta \alpha_i := \alpha_i -\alpha_{i-1}$ represents the angular difference between two adjacent rigid units, and $\alpha_0 =\num{0}$. The sole active torque is applied to the first rigid unit. All specific parameters used in this experiment are summarized in \tabref{tab:parameters}. The following experiments are implemented in \verb|Matlab| and \verb|Simulink|.
\begin{table}[h]
\caption{Parameters of the discrete beam model.}\label{tab:parameters}
\begin{tabular*}{\textwidth}{@{\extracolsep\fill}cccc}
\toprule%
Parameter & Value & Unit & Description \\
\midrule
$n$       & $50$         & -                 & number of rigid units \\
$l_i$     &$\num{3e-2}$  & $\unit{\meter}$   & length of each rigid unit \\
$I_A$     & $\num{5e-3}$ & $\unit{\meter^4}$ & the moment of inertia \\
$k_{i1}$  & $\num{5}$    & \unit{N.m/rad}    & spring coefficient \\
$k_{i2}$  & $\num{1e3}$  & \unit{N.m/rad^3}  & spring coefficient \\
$k_{i3}$  & $\num{1e4}$  & \unit{N.m/rad^5}  & spring coefficient\\
$b_i$     & $30$         & \unit{N.s/rad}    & damper coefficient \\
\botrule
\end{tabular*}
\end{table}

\subsubsection{Reference Trajectory Distribution}
The aim of this experiment is to utilize \algoref{algo:online_quasi_newton} to learn the parameterized networks $\pi_{\text{ff}}$ and $\pi_{\text{fb}}$ in an online manner (see~\figref{fig:control_loop}). The outputs of the parameterized networks yield the active torque $\tau$. The aim of the online learning is to find the parameters $\omega_{\text{ff}}$ and $\omega_{\text{fb}}$, in order to minimize the tracking error of the end-effector ($\left|y-y_{\text{ref}}\right|$) for reference trajectories sampled from the unknown distribution $p_{y_{\text{ref}}}$. The output $y$ describes the distance in $y$-direction between the tip of the beam and the horizontal plane. We observe that the presence of nonlinear spring forces and large deformations, coupled with as many as $100$ hidden states (position and velocity in $y$-direction of each unit), renders this task highly challenging.

All reference trajectories used in the experiment arise from sampling an unknown but fixed distribution. At each iteration, we randomly generate the reference trajectory based on the following principles: \begin{enumerate*}
    \item Over a time span of $T_{\text{sim}}$ seconds, the trajectory starts from rest and eventually returns to its initial position, and remains still for an additional $\qty{0.5}{\second}$.
    \item Apart from the starting and ending points ($y_0$ and $y_T$), two other time points, $t_a$ and $t_b$, will be randomly selected within the time duration $T_{\text{sim}}$. The displacements ($y_a$ and $y_b$) and velocities ($v_a$ and $v_b$) at these moments will also be randomly generated, with the accelerations being set to zero.
    \item The four points are connected using trajectories that minimize jerk\footnote{Jerk is defined as the derivative of acceleration of the third derivative of displacement.}~\citep{OptimalControlEngineering2007, piazzi1997interval}.
\end{enumerate*}
The values of the various parameters are summarized in \tabref{tab:distribution}.
\begin{table}[h]
\caption{Summary of the parameters used for generating reference trajectories.}\label{tab:distribution}
\begin{tabular*}{\textwidth}{@{\extracolsep\fill}ccc}
\toprule%
Parameter & Distribution & Unit \\
\midrule
$t_a$ & Uniform$\left(1.2, 1.8\right)$ & $\unit{\second}$ \\
$t_b$ & Uniform$\left(2.9, 3.5\right)$ & $\unit{\second}$ \\
$y_a$, $y_b$ & Uniform$\left(-0.2, 0.2\right)$ & $\unit{\meter}$\\
$v_a$, $v_b$ & Uniform$\left(-2.0, 2.0\right)$ & $\unit{m/s}$ \\
\botrule
\end{tabular*}
\end{table}

\begin{figure}[ht]
\centering
\includegraphics{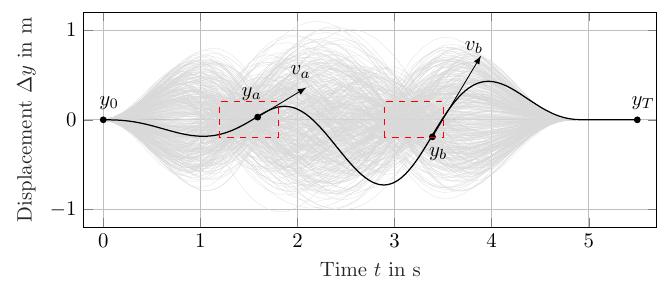}
\caption{The figure illustrates the range of reference trajectories used for training, where the gray lines are composed of $\num{400}$ randomly sampled reference trajectories. The red dashed boxes indicate the spatial and temporal distribution range of the points $y_a$ and $y_b$, respectively.}
\label{fig:dis_traj}
\end{figure}

\figref{fig:dis_traj} illustrates the sampling procedure for the reference trajectories along with $\num{400}$ samples. The total duration is set to $T_{\text{sim}}=\qty{5.5}{\second}$. The red dashed boxes indicate the spatial and temporal distribution range of the points $y_a$ and $y_b$, respectively. We observe that the range of the trajectory is extensive and is not limited to small deformations. In the subsequent experiments, we will see that the parameterized networks trained by our algorithms effectively generalize well across the entire support of $p_{y_{\text{ref}}}$.

\subsubsection{Gradient Estimation}
\begin{figure}[h]
\centering
\begin{subfigure}{0.45\textwidth}
    \centering
    \includegraphics[width=\linewidth]{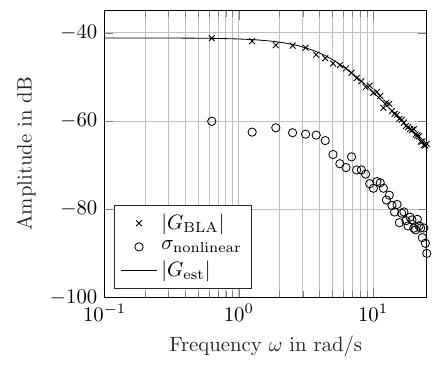}
    \caption{amplitude diagram}
\end{subfigure}
\hskip 0cm
\begin{subfigure}{0.45\textwidth}
    \centering
    \includegraphics[width=\linewidth]{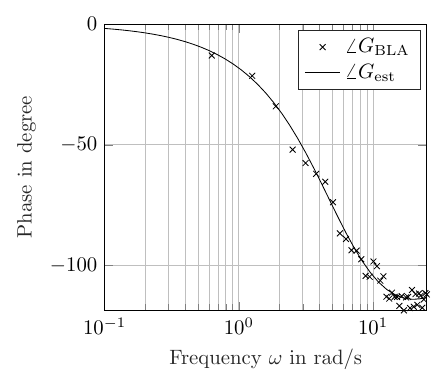}
    \caption{phase diagram}
\end{subfigure}
\caption{The figure displays the amplitude diagram (left) and phase diagram (right) of the system response in the frequency domain. Crosses represent the measured data obtained through system identification in frequency domain, while the solid line represents the fitted transfer function. The nonlinearity system is denoted by circles in the amplitude diagram.}
\label{fig:frequency_response}
\end{figure}

As mentioned in \secref{sec:interpretation}, most of the terms required by \algoref{algo:online_quasi_newton} can be obtained through measurement and computation. However, since the system $G$ is treated as a black-box model, the gradient of the system $\nicefrac{\partial G\left(s_0,u;\zeta\right)}{\partial u}$ cannot be analytically determined. To address this, in this numerical experiment, we employ system identification in the frequency domain to obtain a rough linear estimate of $\nicefrac{\partial G\left(s_0,u;\zeta\right)}{\partial u}$~\citep{pintelonSystemIdentificationFrequency2012a}. We excite the discrete model in \verb|Simulink| with an excitation signal ranging from $\qtyrange{0}{4}{\hertz}$, with an interval of $\qty{0.1}{\hertz}$. The resulting system response in the frequency domain and the estimated linear transfer function are shown in~\figref{fig:frequency_response}. Next, we use the obtained transfer function to construct a linear approximation of $\nicefrac{\partial G\left(s_0,u;\zeta\right)}{\partial u}$, which is denoted by $\mathcal{G}$. It is important to emphasize that in this case the gradient estimate $\mathcal{G}$ is static, meaning that it does not change as a function of $u$. For the specific construction method, please refer to~\citet{maLearningbasedIterativeControl2022b,ReinforcementLearningModelbased}. We observe that the system exhibits a high degree of nonlinearity, which is reflected by the fact that the uncertainty (residual of the linear model) is dominated by the nonlinearities $\sigma_{\text{nonlinear}}$.

Additionally, we estimate the closed-loop system gradient $\mathcal{G}^{\circ}\left(\omega_{\text{ff}}, \omega_{\text{fb}}, \zeta\right)$ using~\eqref{eq:closed_loop3}:
\begin{equation*}
\mathcal{G}^{\circ}\left(\omega_{\text{ff}}, \omega_{\text{fb}}, \zeta\right) = \left(\mathrm{I} - \mathcal{G} \frac{\partial \pi_{\text{fb}}\left(\omega_{\text{fb}};\zeta^{\circ}\right)}{\partial \zeta^{\circ}} \right)^{\dagger} \mathcal{G}.
\end{equation*}

\subsubsection{Network and Input Structure}
\begin{figure}[ht] 
\centering
\begin{subfigure}{0.58\textwidth}
    \centering
    \includegraphics[width=\linewidth]{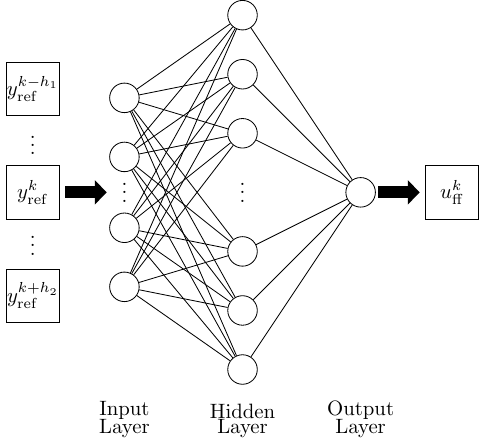}
    \caption{structure of input, output and network for $\pi_{\text{ff}}$}
    \label{fig:pff_beam}
\end{subfigure}
\hskip 0cm
\begin{subfigure}{0.40\textwidth}
    \centering
    \raisebox{1.5cm}{
    \includegraphics[width=\linewidth]{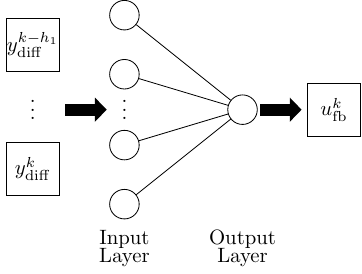}}
    \caption{structure of input, output and network for $\pi_{\text{fb}}$}
    \label{fig:pfb_beam}
\end{subfigure}
\caption{The figure illustrates the input, output, and network structures of both $\pi_{\text{ff}}$ and $\pi_{\text{fb}}$. The feedforward network $\pi_{\text{ff}}$, which is a fully connected network, utilizes the reference trajectory at time $k$, as well as the reference trajectories for the horizons of $h_1$ and $h_2$ before and after this time, as its input. The output is the corresponding feedforward input $u_{k, \text{ff}}$ at time $k$. On the other hand, the feedback network $\pi_{\text{fb}}$, a linear network, employs the trajectory difference for a horizon of $h_1$ units leading up to time $k$ as its input. The output is the respective feedback input $u_{k, \text{fb}}$ at time $k$. The trajectory difference is defined as the difference between the output trajectory and the reference trajectory.}
\label{fig:reconstruction}
\end{figure}

As mentioned in \secref{sec:connection}, each iteration necessitates the cyber-physical system to track a distinct reference trajectory. This means that the parameterized network $\pi$, particularly the feedforward network $\pi_{\text{ff}}$, must be able to adapt to reference trajectories of different lengths. The situation in the case of the beam experiment is illustrated in \figref{fig:reconstruction}. The policy network $\pi_{\text{ff}}$ takes in a horizon of $h_1$ steps in the past and $h_2$ steps in the future to produce the input $u_{k, \text{ff}}$ at time $k$ (see \figref{fig:pff_beam}), while $\pi_{\text{fb}}$ takes only $h_1$ steps in the past to produce $u_{k, {\text{fb}}}$(see \figref{fig:pfb_beam}). In instances where the horizon surpasses the range of the reference trajectory, we employ a zero-padding strategy to compensate for the absent elements.

\subsubsection{Experiments}
In this section, we employ two strategies to parameterize the feedforward network $\pi_{\text{ff}}$. One is a linear network, denoted as $\pi^{1}_{\text{ff}}$. The other is a nonlinear network, represented by $\pi^2_{\text{ff}}$, which is a fully-connected network with a single hidden layer. The \verb|ReLU| function is used as the activation function for the hidden layer, and no activation function is applied to the output layer. We consistently use a linear feedback network $\pi_{\text{fb}}$, and considering the causality of the system, the feedback network only contains the historical information, implying that $h_2 = 0$.

The overview of different experiments is presented in~\tabref{tab:experiments}. In a noise-free environment, trajectory tracking can be viewed as a purely feedforward control task. Therefore, in Experiments $1$-$4$, we employ only the feedforward network $\pi_{\text{ff}}$ and adjust the parameter $\epsilon$, allowing \algoref{algo:online_quasi_newton} to transition between gradient descent ($\epsilon \rightarrow \infty$) and the quasi-Newton method. Through these experiments, we explore the convergence rates of different networks and investigate the influence of different algorithms on convergence as well as their robustness to the selection of hyper-parameters. Subsequently, we intentionally introduce noise $n_{\text{d}}$ to the inputs of the system (see \figref{fig:two_degree}), rendering the pure feedforward network ineffective for the task at hand (see Experiment~$5$). Experiment $6$ demonstrates the ability of the combined feedforward and feedback control ($\pi_{\text{ff}}$ and $\pi_{\text{fb}}$) to resist noise in online learning. 
\begin{table}[h]
\caption{Overview of parameters, network configurations, and experimental results.}\label{tab:experiments}
\begin{tabular*}{\textwidth}{@{\extracolsep\fill}cccccccccc}
\toprule%
\multirow{2}{*}{{No.}} & \multirow{2}{*}{{Model(s)}} & \multirow{2}{*}{$h_1$} & \multirow{2}{*}{$h_2$} &\multirow{2}{*}{{Hidden Neurons}} & \multirow{2}{*}{{Noise}} & \multirow{2}{*}{$\epsilon$} & \multirow{2}{*}{$\alpha$} & \multirow{2}{*}{$\eta$}  & \multirow{2}{*}{{Average Loss}} \\
 & & & & & & & &  & \\
\midrule
 \multirow{2}{*}{$1$} & \multirow{2}{*}{$\pi^1_{\text{ff}}$} &\multirow{2}{*}{$100$} & \multirow{2}{*}{$100$}  &\multirow{2}{*}{-}   &\multirow{2}{*}{no}   &\multirow{2}{*}{$+\infty$}  &\multirow{2}{*}{-}  &\multirow{2}{*}{$0.1$} &\multirow{2}{*}{ $\num{6.90e-3}$} \\
  &  &  &  &  & & & & &\\
    \multirow{2}{*}{$2$} & \multirow{2}{*}{$\pi^1_{\text{ff}}$} &\multirow{2}{*}{$100$} & \multirow{2}{*}{$100$}  &\multirow{2}{*}{-}   &\multirow{2}{*}{no}   &\multirow{2}{*}{$1.0$}  &\multirow{2}{*}{$0.1$}  &\multirow{2}{*}{$15.0$} &\multirow{2}{*}{$\num{6.30e-3}$} \\
  &  &  &  &   & & & & &\\
    \multirow{2}{*}{$3$} & \multirow{2}{*}{$\pi^2_{\text{ff}}$} &\multirow{2}{*}{$100$} & \multirow{2}{*}{$100$}  &\multirow{2}{*}{$40$}   &\multirow{2}{*}{no}   &\multirow{2}{*}{$+\infty$}  &\multirow{2}{*}{-}  &\multirow{2}{*}{$0.1$} &\multirow{2}{*}{$\num{8.27e-4}$}\\
  &  &  &  &  & & & & &\\
   \multirow{2}{*}{$4$} & \multirow{2}{*}{$\pi^2_{\text{ff}}$} &\multirow{2}{*}{$100$} & \multirow{2}{*}{$100$}  &\multirow{2}{*}{$40$}   &\multirow{2}{*}{no}   &\multirow{2}{*}{$1.0$}  &\multirow{2}{*}{$0.1$}  &\multirow{2}{*}{$15.0$} &\multirow{2}{*}{$\num{3.19e-4}$}\\
  &  &  &  &  & & & & &\\
   \multirow{2}{*}{$5$} & \multirow{2}{*}{$\pi^2_{\text{ff}}$} &\multirow{2}{*}{$100$} & \multirow{2}{*}{$100$}  &\multirow{2}{*}{$40$}   &\multirow{2}{*}{yes}   &\multirow{2}{*}{$1.0$}  &\multirow{2}{*}{$0.1$}  &\multirow{2}{*}{$15.0$} &\multirow{2}{*}{$\num{1.10e-3}$}\\
  &  &  &  &  & & & & &\\
    \multirow{2}{*}{$6$} &$\pi_{\text{fb}}$ & $25$ & -  &-    &\multirow{2}{*}{yes}   &\multirow{2}{*}{$1.0$}  &\multirow{2}{*}{$0.1$}  &\multirow{2}{*}{$15.0$} &\multirow{2}{*}{$\num{4.65e-4}$}\\
    &$\pi_{\text{ff}}^2$ & $100$ &$100$  &$40$  &  & & & & \\
\botrule
\end{tabular*}
\end{table}

\begin{figure}[ht] 
\centering
\begin{subfigure}{0.32\textwidth}
    \centering
    \includegraphics[width=\linewidth]{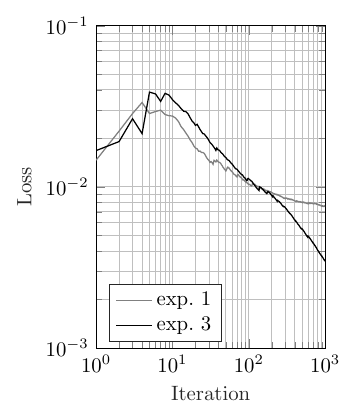}
    \caption{results of exp.~1 and 3}
    \label{fig:beam_loss1}
\end{subfigure}
\hskip 0cm
\begin{subfigure}{0.32\textwidth}
    \centering
    \includegraphics[width=\linewidth]{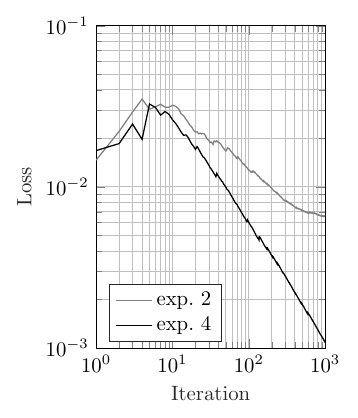}
    \caption{results of exp.~2 and 4}
    \label{fig:beam_loss2}
\end{subfigure}
\hskip 0cm
\begin{subfigure}{0.32\textwidth}
    \centering
    \includegraphics[width=\linewidth]{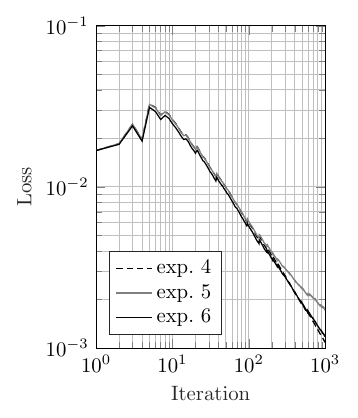}
    \caption{results of exp.~4, 5 and 6}
    \label{fig:beam_loss3}
\end{subfigure}
\caption{This figure depicts the convergence results $\delta_t,~t=1,\dots,1000$ of different experiments.}
\label{fig:beam_loss}
\end{figure}

In each experiment, we train the parameterized networks for $\num{1000}$ iterations. The average loss $\delta_t$ is given by 
\begin{equation*}
    \delta_t = \frac{1}{t} \sum_{k=1}^{t} l\left(y_k;y_{\text{ref},k}\right),~t=1,\dots,T.
\end{equation*}
The convergence results of Experiments $1$ and $3$ are illustrated in~\figref{fig:beam_loss1}, while the results of Experiments $2$ and $4$ are shown in~\figref{fig:beam_loss2}. We note that, limited by the complexity of the model, the convergence results of nonlinear models are significantly better than those of linear models, regardless of whether the gradient descent algorithm or the quasi-Newton method is used. At the same time, we also notice that, under the premise of using nonlinear models, the convergence results of the quasi-Newton method are superior to those of the gradient descent algorithm. We believe that this is due to Term~$2$ in~\eqref{eq:trust_region}, which is used to capture the local curvature of the loss functions, and the absence of Term~$1$ in~\eqref{eq:trust_region}, leading to a lack of constraint on the changes in the outputs of $\pi_{\text{ff}}$ and $\pi_{\text{fb}}$. In our experiments, such a constraint is important for our online learning tasks, ensuring that the outputs between iterations do not change drastically, thereby preventing oscillations. Moreover, the quasi-Newton method demonstrates greater robustness in the adjustment of hyperparameters, such as the step size $\eta$, which requires additional fine-tuning for the gradient descent algorithm.  

In Experiment $5$, we artificially introduce noise $n_{\text{d}}$ to render the purely feedforward control ineffective in trajectory tracking and introduce a feedback controller in Experiment $6$ to reject noise, with the convergence results shown in \figref{fig:beam_loss3}. The introduction of the feedback controller successfully rejects the process noise.

 Additionally, we evaluate the performance of all the obtained parameterized networks trained in different experiments on a newly generated test data set previously unseen by our algorithms (see the average loss in \tabref{tab:experiments}), in order to investigate the generalization capability of the networks. Although we only utilize a linear static gradient estimate $\mathcal{G}$ (which, unsurprisingly, is a very poor estimate), the algorithms still perform well using either gradient descent method or quasi-Newton method, reflecting its strong robustness to modeling errors.

\begin{figure}[ht]
\centering
\begin{subfigure}{0.48\textwidth}
    \centering
    \includegraphics[width=\linewidth]{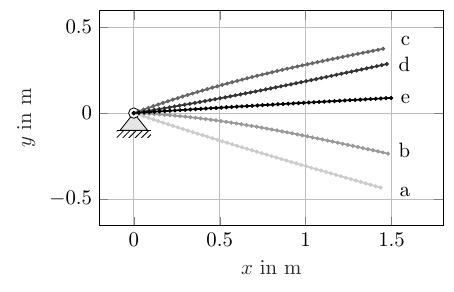}
    \caption{motion of the beam in exp.~$2$}
\end{subfigure}
\hskip 0cm
\begin{subfigure}{0.48\textwidth}
    \centering
    \includegraphics[width=\linewidth]{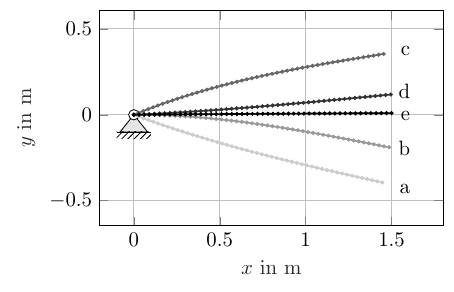}
    \caption{motion of the beam in exp.~$4$}
\end{subfigure}
\vskip 0.5cm
\begin{subfigure}{0.8\textwidth}
    \centering
    \includegraphics[width=\linewidth]{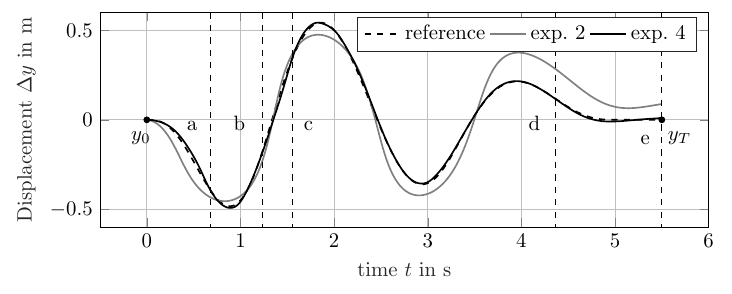}
    \caption{motion of the tip in $y$-direction}
\end{subfigure}
\caption{This figure shows the overall motion of the beam (upper) and the motion of the tip in the $y$-direction (lower) when the models obtained in Experiments~$2$ and~$4$ track the reference trajectory from test data set. The motion starts from a standstill and proceeds in the order of a-b-c-d-e, with the corresponding moments marked with the same labels in the lower subfigure.}
\label{fig:beam_motion}
\end{figure}

Lastly, we select a trajectory from the test data set to demonstrate the tracking performance of the models obtained in Experiments $2$ and $4$. The upper subfigure of~\figref{fig:beam_motion} shows the positions of the beam controlled by the linear and nonlinear models at different selected moments. The lower subfigure illustrates the tracking error of the beam tip in the $y$-direction as controlled by the different models.

\subsubsection{Benchmarks with other Approaches}
\begin{figure}[ht]
\centering
\begin{subfigure}{0.426\textwidth}
    \centering
    \includegraphics[width=\linewidth]{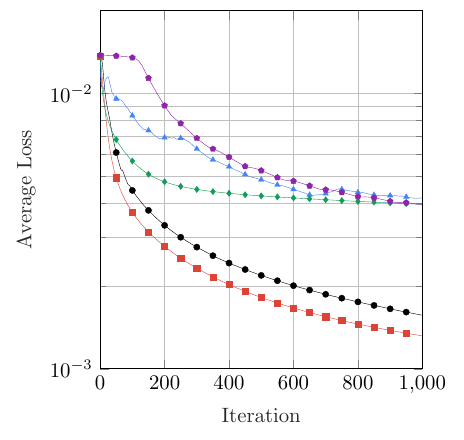}
    \caption{tracking single trajectory}
    \label{fig:comp_benchmark_a}
\end{subfigure}
\hskip 0cm
\begin{subfigure}{0.534\textwidth}
    \centering
    \includegraphics[width=\linewidth]{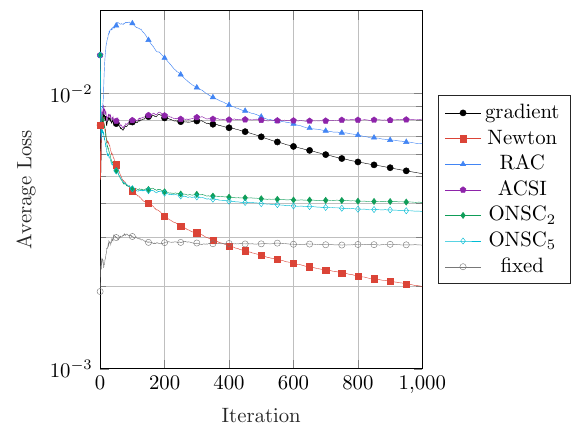}
    \caption{tracking multiple trajectories}
    \label{fig:comp_benchmark_b}
\end{subfigure}
\caption{The figure presents the trajectory tracking results of different algorithms. The left plot depicts tracking performance for a single fixed trajectory, whereas the right plot shows the performance across multiple randomly sampled trajectories.}
\label{fig:comp_benchmark}
\end{figure}

\begin{editenv}
    We compare the performance of several algorithms from \secref{sec:related_work} with our proposed methods in this experiment. The results are shown in \figref{fig:comp_benchmark}. Specifically, the Robust Adaptive Control (RAC) algorithm is based on the work of \citet{deanRegretBoundsRobust2018}, the Adversarial Control via System Identification (ACSI) algorithm is based on the work of \citet{hazanNonstochasticControlProblem2020,agarwalOnlineControlAdversarial2019}, and the Online Non-Stochastic Control (ONSC) algorithm is based on the work of \citet{yanOnlineNonstochasticControl} where the different subscripts indicate the number of random perturbations used for estimating the linear model.
\end{editenv}
\begin{editenv}
\figref{fig:comp_benchmark_a} illustrates the performance of different algorithms on the same fixed trajectory. We observe that all algorithms eventually converge, but our methods exhibit both faster convergence rates and better final performance. \figref{fig:comp_benchmark_b} shows the performance under a fixed number of experiments. We find that only our quasi-Newton method outperforms the fixed best linear controller. The ACSI algorithm does not adapt well to this setting and shows no signs of convergence. For the ONSC algorithm, increasing the number of perturbations used for model estimation slightly improves performance, but also significantly increases computational complexity.
\end{editenv}

\subsection{Four-Legged Robot}
\label{sec:ant}
In this simulation, we adopt the ant model~\citep{schulmanHighDimensionalContinuousControl2018a} frequently used to demonstrate RL algorithms to evaluate the effectiveness of our algorithms. The ant model is a quadruped robot with four legs symmetrically distributed around the torso, each connected by two hinged joints, providing two degrees of rotational freedom per leg and eight degrees in total for the ant. In our simulation environment, we have chosen \verb|Isaac Gym|~\citep{makoviychuk2021isaac} for its ability to support large-scale parallel simulations, which enables us to rapidly estimate system gradients using a stochastic finite difference method. It is important to emphasize that the traditional RL task on this model focuses on enabling the ant to move forward as quickly as possible. However, in our experiment, our aim is to enable the ant to track any reference trajectory of the center of the mass of the torso. It should be noted that in this experiment, we do not artificially introduce system noise, thus the trajectory tracking task can be considered a purely feedforward control task. Therefore, we only employ a feedforward model $\pi_{\text{ff}}$, which implies that $\pi_{\text{fb}} = 0$.

We recognize that the learning of the motion of the ant, without any prior knowledge, is a challenging task. Compared to the numerical examples mentioned in~\secref{sec:numerical_experiment}, the learning of the motion present the following differences and difficulties: First, due to the contacts and interactions between the ant and the environment, the motion of the ant is non-smooth, and accordingly, its gradients are discontinuous (though still assumed to be bounded). Second, the states of the ant are not fully observable. In fact, in this experiment only the information about the torso is assumed to be measurable and observable, including its positions, orientations, and corresponding velocities and angular velocities. This means that changes in control inputs do not necessarily cause changes in the outputs. For instance, when one of the legs is not in contact with the ground, the positional change of this leg caused by input variation will not affect the posture of the torso. Third, walking, as a periodic behavior, should follow specific gaits and frequencies. Training a network model from scratch may lead to the ant exhibiting anomalous behaviors.

To overcome the challenges mentioned above, we make the following adjustments to the networks in this experiment:
\begin{enumerate}
    \item We input the entire reference trajectory for $k=0$ to $k=T_{\text{sim}}$ into the model and predict the entire control sequence for $k=0$ to $k=T_{\text{sim}}$.
    \item We fix the duration of the trajectory, but we can still adapt the networks to varying trajectory lengths through appropriate preprocessing.
    \item We use a pre-trained linear model to provide prior knowledge of ant motion patterns.
\end{enumerate}

\subsubsection{Reference Trajectory Distribution}
In this experiment, we can only measure and observe the information about the torso, which includes the position of the torso, its orientation represented by a quaternion, and the translational and angular velocities of the torso. The ant moves on a rough and infinitely flat plane, therefore the reference trajectories contain only three components: the planar position of the torso, i.e., the $x$ and $y$ coordinates, and the yaw, which is the rotation of the torso around the $z$-axis. 

\begin{figure}[!h]
\centering
\begin{subfigure}{0.50\textwidth}
    \centering
    \includegraphics[width=\linewidth]{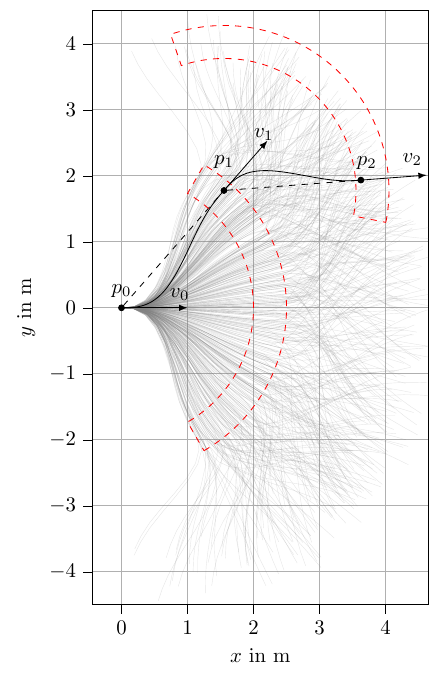}
    \caption{distribution of reference trajectories}
\end{subfigure}
\hskip 0cm
\begin{subfigure}{0.48\textwidth}
    \centering
    \includegraphics[width=\linewidth]{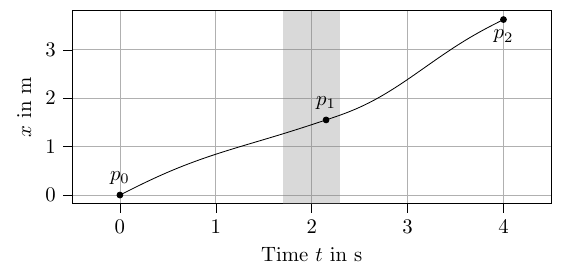}
    \caption{trajectory of $x$}
    \includegraphics[width=\linewidth]{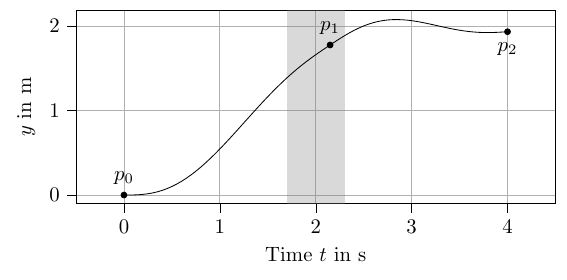}
    \caption{trajectory of $y$}
    \includegraphics[width=\linewidth]{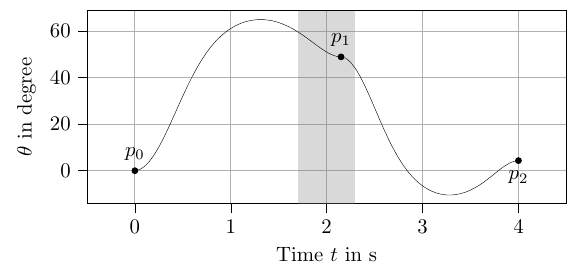}
    \caption{trajectory of yaw}
\end{subfigure}
\caption{The left subfigure shows the area generated by $\num{500}$ randomly sampled reference trajectories used for tracking (depicted as grey lines), along with an example reference trajectory to illustrate the rules for generating trajectories (shown as a black line). The red dashed disks represent the distribution range for the position of the next point in the $x$-$y$ plane, assuming the previous point is determined. The radial gap of the disk is $\qty{0.5}{\meter}$, with an angular span $\pm$ $\qty{60}{\degree}$ centered around the tangent direction at the previous point. The right subfigure illustrates the temporal evolution of the example reference trajectory, showing its $x$, $y$, and yaw components. The time points for generating points $p_0$ and $p_2$ are fixed. The grey areas in the right subfigures represent the time  range for point $p_1$, which is centered around $t = \qty{2}{\second}$ with a permissible deviation of $\pm \qty{0.3}{\second}$.}
\label{fig:ant_reference}
\end{figure}

\figref{fig:ant_reference} displays the distribution of the reference trajectories used for tracking. We take one of the sampled trajectories as an example to illustrate the general rules for generating reference trajectories. The trajectories are generated over a time duration of $T_{\text{sim}}=\qty{4}{\second}$. The starting point $p_0$ is fixed at the point $\left[0,0\right]^{\text{T}}$ in the $x$-$y$ plane at time $t_0=\qty{0}{\second}$, and the initial velocity $v_0$ is also fixed at $\qty{1}{\meter \per \second}$ directed along the positive $x$-axis. Next, we uniformly generate the point $p_1$ within a disk centered around $p_0$ with radii of $\qty{2}{\meter}$ and $\qty{2.5}{\meter}$, and an angular span of $\pm \qty{60}{\degree}$ centered around $v_0$ (see the red dashed disk in the left subfigure). The velocity $v_1$ at $p_1$ is also set to $\qty{1}{\meter \per \second}$, in the direction of the line from $p_0$ to $p_1$. The time $t_1$ for generating $p_1$ is uniformly within a range of $\pm \qty{0.3}{\second}$ centered around $t=\qty{2}{\second}$ (see the shadow areas in the right subfigures). Based on the point $p_1$, the point $p_2$ and its corresponding velocity $v_2$ are generated in the same manner, with the time point $t_2=\qty{4}{\second}$ being fixed for $p_2$. The acceleration at each point is set to zero. Finally, we connect these three points using a trajectory that minimizes jerk. We note that the duration of all trajectories is fixed. However, since the time duration $T_{\text{sim}}$ is sufficiently long to accommodate multiple gaits, trajectories with varying durations can still be accommodated simply by periodically reapplying $\pi_{\text{ff}}$ and $\pi_{\text{fb}}$.

\subsubsection{Gradient Estimation}
In the context of the ant model, which is a system characterized by contacts and non-smooth motion, employing system identification methods as described in \secref{sec:numerical_experiment} is not applicable. Fortunately, the powerful parallel simulation capability of the \verb|Isaac Gym| environment allows us to easily estimate the system gradient $\mathcal{G} \left(u_t\right)$ using a stochastic finite difference method. At each iteration, we run $n_{\text{env}}$ (here $n_{\text{env}}=2000$) identical environments in parallel in addition to the nominal environment. The nominal input $u_t,~t=1,\dots,T$, is fed into the nominal environment, yielding the corresponding nominal output $y_t$. For the remaining parallel environments, normally distributed noise $n_{\text{d}}$ with a mean of zero and a variance of one is added to the nominal input $u_t$ (see~\figref{fig:two_degree}), denoted as $\widetilde{u}_{t,i},~i=1,\dots,n_{\text{env}}$, resulting in the respective outputs $\widetilde{y}_{t,i}$. Finally, we estimate the system gradient $\mathcal{G} \left(u_t\right)$ using least squares\footnote{In the experiment, only the ants that remain upright until the end are considered for estimating the gradient.}:
\begin{equation*}
    \mathcal{G} \left(u_t\right)^{\text{T}} = 
    \begin{bmatrix}
    \left(\widetilde{u}_{t, 1} - u_t\right)^{\text{T}} \\
    \left(\widetilde{u}_{t, 2} - u_t\right)^{\text{T}}\\
    \vdots\\
    \left(\widetilde{u}_{t, n_{\text{env}}} - u_t\right)^{\text{T}}
    \end{bmatrix}^{\dagger}  
    \begin{bmatrix}
    \left(\widetilde{y}_{t, 1} - y_t\right)^{\text{T}} \\
    \left(\widetilde{y}_{t, 2} - y_t\right)^{\text{T}}\\
    \vdots\\
    \left(\widetilde{y}_{t,  n_{\text{env}}} - y_t\right)^{\text{T}}
    \end{bmatrix},
\end{equation*}
where we stack all inputs and outputs by columns respectively.

\subsubsection{Neural Network with Pre-Trained Motion Patterns}
In order to enable online learning with the ant model, we parameterize our networks as follows:
\begin{equation*}
    \pi_{\text{ff}}\left(\omega_{\text{ff}}; \zeta \right) = U  \phi \left(\omega_{\text{ff}};  V^{\text{T}} \zeta \right),~\pi_{\text{fb}} = 0,
\end{equation*}
where the matrices $U \in \mathbb{R}^{mq \times n_{\sigma}}$ and $V \in \mathbb{R}^{nq \times n_{\sigma}}$ represent linear transformation to a lower dimensional latent space and $\phi: \mathbb{R}^{n_{\omega_{\text{ff}}}} \times \mathbb{R}^{n_{\sigma}} \rightarrow \mathbb{R}^{mq}$ is a neural network comprising one hidden layer.
The matrices $U$ and $V$ are obtained through a pretraining process based on $\num{45}$ trajectories. Further details about the pretraining process and obtaining the matrices $U$ and $V$ are included in Appendix~\ref{sec:AppendixB}.

\subsubsection{Experiments}
\begin{figure}[!h]
\centering
\includegraphics{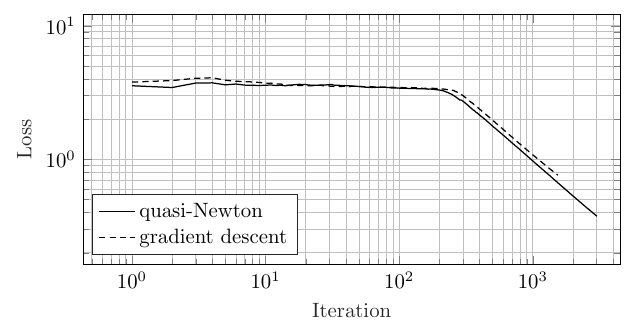}
\caption{The figure shows the convergence results of the gradient descent and quasi-Newton method. The gray line represents the average loss of the gradient descent method, and the black line indicates the average loss of the quasi-Newton method.}
\label{fig:ant_loss}
\end{figure}

In this experiment, we use a fully connected network with only one hidden layer containing $\num{20}$ neurons. The hidden layer employs the \verb|ReLU| activation function, while the output layer does not have an activation function. We employ different methods to train the network, and the parameters are shown in \tabref{tab:experiments_ant}.
\begin{table}[h]
\caption{Parameters for training the ant model.}\label{tab:experiments_ant}
\begin{tabular*}{\textwidth}{@{\extracolsep\fill}ccccccc}
\toprule%
No. & Model & $n_{\sigma}$ & Hidden Neurons & $\epsilon$ & $\alpha$ & $\eta$ \\
\midrule
$1$  & $\pi_{\text{ff}}$ & $90$ & $45$ & $+\infty$  & - & Diminishing \\
$2$  & $\pi_{\text{ff}}$ & $45$ & $20$ & $0.1$  & $0.5$ & $0.1$ \\
\botrule
\end{tabular*}
\end{table}
The two experiments were conducted with over $\num{1500}$ and $\num{3500}$ iterations, respectively, and their convergence results are shown in~\figref{fig:ant_loss}. We note that the loss of both algorithms eventually converges to the same level with the same rate. However, it is important to emphasize that to ensure the convergence of the gradient descent method, its step size $\eta_t$ must be carefully designed. In contrast, the quasi-Newton method demonstrates much stronger robustness to the step size selection. Finally, through this experiment, we demonstrate that in such complex cyber-physical system, even with a poor gradient estimate, our algorithms still ensure convergence and exhibit high robustness to modeling errors. It is important to emphasize that, unlike RL, which optimizes a feedback policy to enable the ant to move forward as fast as possible, our algorithms learn a feedforward model $\pi_{\text{ff}}$ that allows the ant to track any reference trajectories sampled from the distribution $p_{y_{\text{ref}}}$, thereby truly enabling it to learn the skill of walking. Additionally, our algorithms are capable of continuously improving the tracking performance of the feedforward network through online learning during deployment.

\begin{figure}[!h]
\centering
\begin{subfigure}{0.50\textwidth}
    \centering
    \includegraphics[width=\linewidth]{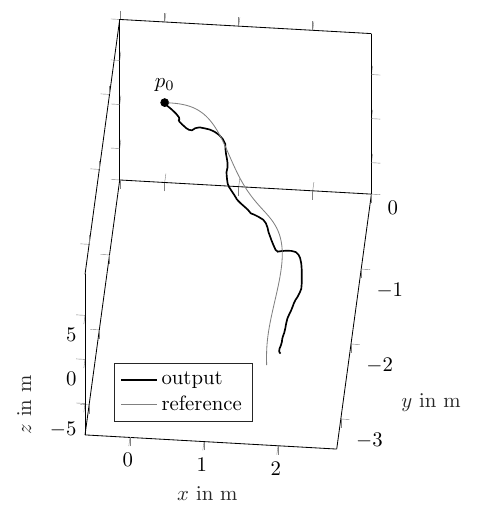}
    \caption{trajectory in the three-dimensional space}
\end{subfigure}
\hskip 0cm
\begin{subfigure}{0.48\textwidth}
    \centering
    \includegraphics[width=\linewidth]{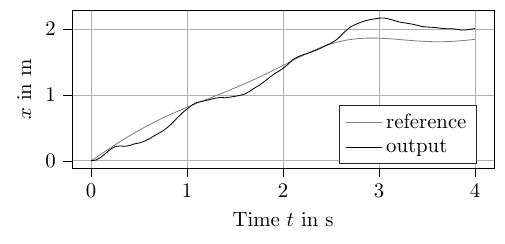}
    \caption{trajectory of $x$}
    \includegraphics[width=\linewidth]{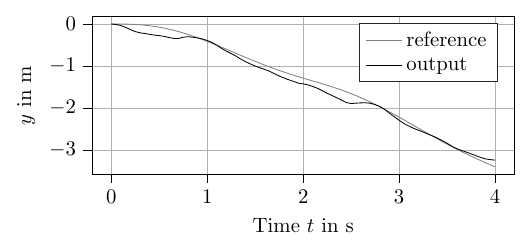}
    \caption{trajectory of $y$}
    \includegraphics[width=\linewidth]{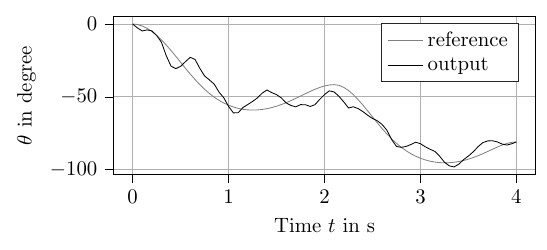}
    \caption{trajectory of yaw}
\end{subfigure}
\caption{The figure shows the tracking performance of the network trained by the quasi-Newton method on a single reference trajectory. The left subfigure illustrates the tracking effect of the network in the three-dimensional space, where the $z$-value of the reference trajectory is chosen to be $\qty{0.5}{\meter}$. The right subfigures separately show the tracking performance of the network for the $x$, $y$, and yaw components.}
\label{fig:ant_ol}
\end{figure}

We select one of the trajectories in~\figref{fig:ant_ol} to demonstrate the tracking performance of the neural network trained by the quasi-Newton method. The left subfigure shows the tracking performance in the three-dimensional space, where the reference trajectory is set at a fixed height of $\qty{0.5}{\meter}$. The right subfigures separately demonstrate the tracking performance of the model for the $x$, $y$, and yaw components of the trajectory.

\subsection{Table Tennis Robot}
\begin{figure}[ht]
\centering
\includegraphics[width=0.5\linewidth]{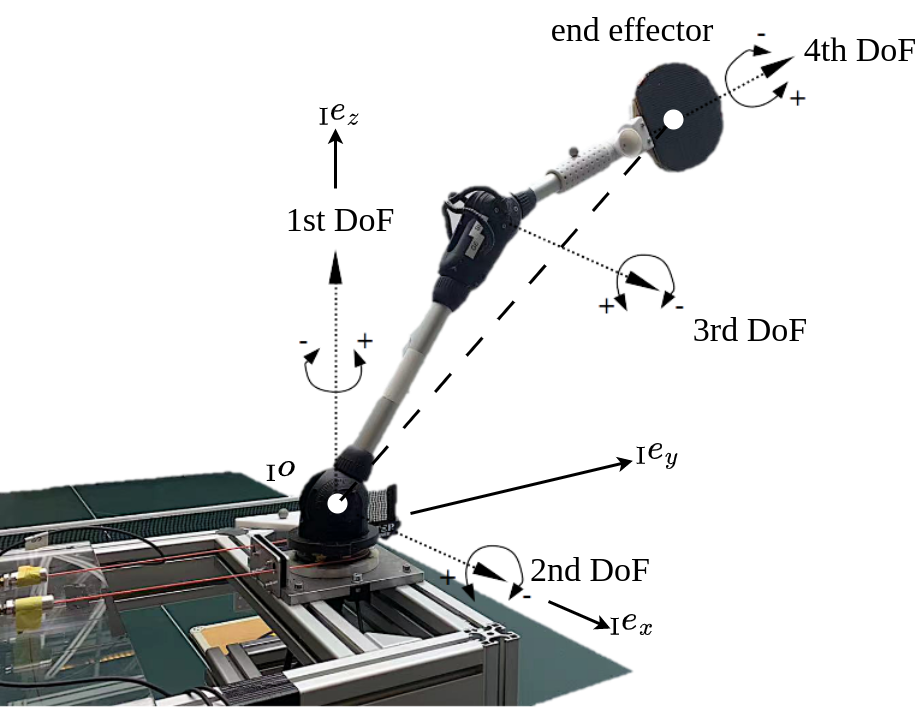}
\caption{The figure shows the structure of the robot arm. It has four rotational joints, and each joint is actuated by a pair of PAMs. The unit vectors $_{\mathrm{I}}e_{x}$, $_{\mathrm{I}}e_{y}$ and $_{\mathrm{I}}e_{z}$ together form the global coordinate system $\left\{\mathrm{I}\right\}$ with $_{\mathrm{I}}o$ as the origin. For simplicity we consider only DoF $1$-$3$, whereas DoF $4$ is controlled with a PID controller.}
\label{fig:pamy}
\end{figure}

In this experiment, we evaluate the effectiveness of our algorithms in a real-world cyber-physical system, which is a robotic arm actuated by pneumatic artificial muscles (PAMs) as shown in~\figref{fig:pamy}. This robotic arm is designed for playing table tennis. The PAMs offer a high power-to-weight ratio enabling rapid movements, but also introduce substantial nonlinearities to the system, making the precise control of this robotic arm particularly challenging. 

The robot has four rotational degrees of freedom (DoF), each powered by a pair of PAMs. For more information about this robot, please refer to~\citet{maLearningbasedIterativeControl2022b, buchlerLearningPlayTable2022a, buchlerLightweightRoboticArm2016b}. In this experiment, we only consider learning the controller for the first three degrees of freedom, as the last degree of freedom can be accurately controlled with a PID controller.

\subsubsection{Reference Trajectory Distribution}
In this experiment, the distribution of the reference trajectories is tailored to the specific task of playing table tennis. First, we fix the initial posture $p_{\text{ini}}$ of the robot, and then determine the interception point $p_{\text{int}}$ based on the incoming trajectory of the table tennis ball. The robot moves from the initial posture to the interception point and then returns to the initial posture. Each trajectory is planned within a polar coordinate system using a minimum jerk trajectory that connects $p_{\text{ini}}$ and $p_{\text{int}}$. For more detailed information and the specific form of the trajectory, please refer to~\citet{maLearningbasedIterativeControl2022b, ReinforcementLearningModelbased}. 

\subsubsection{Gradient Estimation}
Each degree of freedom is highly coupled with the others. However, for estimating the gradient, we disregard the coupling between each degree of freedom, thereby decomposing the system into three independent subsystems. This allows for the gradient estimation for each degree of freedom separately. The input for each degree of freedom consists of the pressure values from two artificial muscles, and the output is the angle of the degree of freedom. We further reduce the number of inputs by setting a baseline pressure and controlling the pressure difference between the two artificial muscles and the baseline. Subsequently, we use the method described in~\secref{sec:numerical_experiment} to identify the transfer function for each simplified degree of freedom in the frequency domain and use this to estimate a static gradient $\mathcal{G}^i,~i=1,2,3$ for each degree of freedom. Please refer to~\citet{maLearningbasedIterativeControl2022b} for details on the system identification and the construction of the approximate gradient.

\subsubsection{Network and Input Structure}
We construct three structurally identical linear network models, $\pi_{\text{ff}}^{i},~i=1,2,3$ as in~\secref{sec:numerical_experiment}. Although the coupling between different degrees of freedom is ignored for the gradient estimation, the coupling is accounted for by the online learning. Therefore, we use the reference trajectories of all three degrees of freedom as inputs to each linear model. The method for generating the inputs is also described in~\secref{sec:numerical_experiment}, where $h = 100$.

\subsubsection{Experiments}
The reference trajectories are derived from past interceptions of table tennis balls, and we use $\num{30}$ reference trajectories for training. In our online learning approach, we randomly sample a reference trajectory without replacement. Once every reference trajectory has been executed we start a new epoch, where we again sample one trajectory at random without replacement.

\begin{figure}[ht]
\centering
\includegraphics{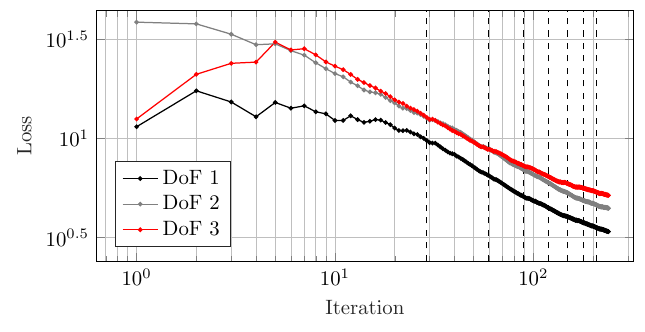}
\caption{This figure illustrates the average tracking error for each degree of freedom. Each independent linear model is trained by a pseudo-online learning approach, wherein each trajectory is randomly sampled without replacement in every epoch. A total of $\num{30}$ reference trajectories are available for training in each epoch. Each epoch is distinguished in the figure by vertical dashed lines.}
\label{fig:pamy_loss}
\end{figure}

The convergence of the online learning is shown in \figref{fig:pamy_loss}, with each epoch distinguished by vertical dashed lines. It is observed that the average tracking errors of all three linear models converge with the same rate. This suggests that through the training, the linear models are capable of capturing the coupling effects between different degrees of freedom. It is important to emphasize that even with a rough estimate of the gradient, our algorithm achieves rapid convergence and maintains stability in a highly complex nonlinear system. This not only underlines the robustness of our algorithm against modeling errors but also provides empirical evidence of its performance on real-world systems.

\begin{figure}[ht]
\centering
\begin{subfigure}{0.50\textwidth}
    \centering
    \includegraphics[width=\linewidth]{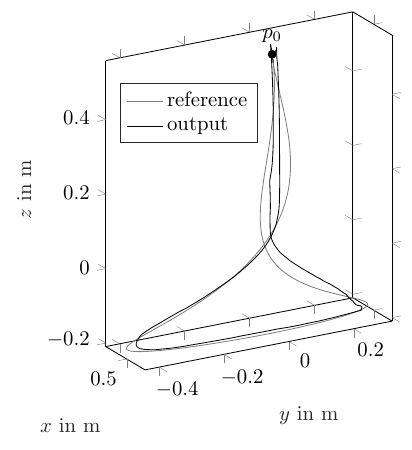}
    \caption{trajectory in three-dimensional space}
\end{subfigure}
\hskip 0cm
\begin{subfigure}{0.48\textwidth}
    \centering
    \includegraphics[width=\linewidth]{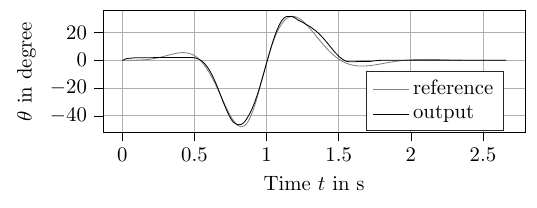}
    \caption{trajectory of DoF $1$}
    \includegraphics[width=\linewidth]{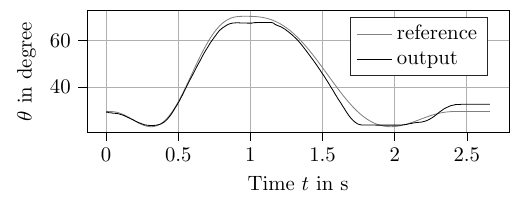}
    \caption{trajectory of DoF $2$}
    \includegraphics[width=\linewidth]{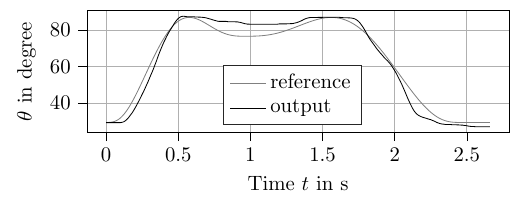}
    \caption{trajectory of DoF $3$}
\end{subfigure}
\caption{The figure demonstrates the tracking performance of a reference trajectory from the last training epoch. The left subfigure shows the tracking performance of the tip of the robot in three-dimensional space. The right subfigures separately show the tracking performance for each degree of freedom.}
\label{fig:pamy_ol}
\end{figure}

We select one trajectory from the last training epoch to demonstrate the tracking performance of the linear models. In~\figref{fig:pamy_ol}, the left subfigure shows the tracking performance of the tip of the robot in three-dimensional space. The right subfigures separately show the tracking performance for each degree of freedom.

\section{Conclusions}
\label{sec:conclusion}
In this article, we propose a novel gradient-based online learning framework derived from a trust-region approach, operating under the assumptions that the loss functions are smooth but not required to be convex. Thanks to gradient information that is incorporated within the algorithm, we obtain a sample efficient online learning approach that is applicable to cyber-physical and robotic systems. The framework presented in this article includes a stochastic optimization algorithm, various designs for neural networks and input structures for feedforward and feedback control scenarios. We have not only theoretically proven the convergence of the algorithm without relying on convexity, but also evaluated the effectiveness of our proposed framework through a wide range of examples, including numerical experiments, simulation experiments, and real-world implementations. These experiments highlight fast convergence of our algorithms and robustness against modeling errors. Furthermore, they provide empirical evidence that this algorithm can be deployed in real-world applications in the future.

\backmatter

\bmhead{Acknowledgements}
Hao Ma and Michael Muehlebach thank the German Research Foundation, the Branco Weiss Fellowship, administered by ETH Zurich, and the Center for Learning Systems for the funding support.

\section*{Disclosures and Declarations}

\bmhead{Funding}
This work received funding from the German Research Foundation, the Branco Weiss Fellowship, administered by ETH Zurich, and the Center for Learning Systems.

\bmhead{Conflict of interest/Competing interests}
The authors have no competing interests to declare that are relevant to the content of this article.

\bmhead{Ethics approval and consent to participate}
Not applicable.

\bmhead{Consent for publication}
Not applicable.

\bmhead{Data availability}
Not applicable.

\bmhead{Code availability} 
Not applicable.

\bmhead{Author contribution} 
M. Michael and H. Ma conceived the original idea. The theoretical analysis presented in the paper was contributed to by M. Zeilinger, M. Michael and H. Ma. H. Ma performed all the coding and experiments. M. Zeilinger and M. Michael provided experimental concepts and guided the experiments based on the results. The initial draft of the manuscript was jointly prepared by M. Michael and H. Ma. The manuscript was revised based on valuable feedback from M. Zeilinger and was submitted with the joint agreement of M. Zeilinger and M. Michael.


\begin{appendices}

\section{Proof of \thmref{thm:online_quasi_newton}}
\label{sec:appendixA}
In this appendix we will prove~\thmref{thm:online_quasi_newton}. First of all, we show that~\aspref{asp:model_error} implies the following inequality:
\begin{equation*}
    \left| \mathbb{E}_{\zeta} \left[ \left. \mathcal{F} \left(\omega_t; \zeta\right) \right| \omega_t \right] - \mathbb{E}_{\zeta} \left[ \left. \nabla f \left(\omega_t; \zeta\right) \right|  \omega_t \right] \right|^2_{A_{t-1}^{-1}} \leq \kappa^2 \left|\mathbb{E}_{\zeta} \left[ \left. \nabla f \left(\omega; \zeta\right) \right| \omega \right] \right|^2_{A_{t-1}^{-1}}.
\end{equation*}

\begin{lemma}\label{lmm:modeling_error}
    Let
    \begin{equation}
        \lambda_{\mathrm{max}} \left(P\right) \left|y-x\right|^2 \leq \kappa^2 \lambda_{\mathrm{min}} \left(P\right) \left|x\right|^2
    \label{eq:lemma_eq1}
    \end{equation}
    be satisfied for all $x,y \in \mathbb{R}^n$, $\kappa \in \left[0,1\right)$, and for a positive definite matrix $P \in \mathbb{R}^{n\times n}$, then the following inequality holds
    \begin{equation*}
        \left|y - x\right|^2_{P} \leq \kappa^2 \left|x\right|^2_{P}.
    \end{equation*}
\end{lemma}
\begin{proof}
    We prove~\lmmref{lmm:modeling_error} via a contradiction argument. We first assume that there exists $x,y\in\mathbb{R}^n$ such that~\eqref{eq:lemma_eq1} is satisfied and the following inequality holds
    \begin{equation*}
        \left|y-x\right|^2_{P} > \kappa^2 \left|x\right|^2_{P},
    \end{equation*}
    which implies that
    \begin{equation*}
        \lambda_{\mathrm{max}} \left(P\right) \left|y-x\right|^2 \geq \left|y - x\right|^2_{P} > \kappa^2 \left|x\right|^2_{P} \geq \kappa^2 \lambda_{\mathrm{min}} \left(P\right) \left|x\right|^2.
    \end{equation*}
    Then we have
    \begin{equation*}
        \lambda_{\mathrm{max}} \left(P\right) \left|y-x\right|^2 >  \kappa^2 \lambda_{\mathrm{min}} \left(P\right) \left|x\right|^2,
    \end{equation*}
    which contradicts~\eqref{eq:lemma_eq1}. This concludes the proof of~\lmmref{lmm:modeling_error}.
\end{proof}
We note that a key difficulty in the proof of~\thmref{thm:online_quasi_newton} is that the stochastic gradients and the matrices $A_t^{-1}$ are dependent random variables. We will therefore in a first step, represent the inverse of the matrix $A_t$ using $A_{t-1}^{-1}$ in order to decouple the dependency.

\begin{proposition}\label{prop:A_t}
Consider the sequence of pseudo-Hessian matrices $A_t$ obtained according to~\algoref{algo:online_quasi_newton}. Then, the following relationship between $A_{t-1}^{-1}$ and $A_{t}^{-1}$ holds for all $t \geq 2$,
\begin{equation*}
A_{t}^{-1} = \frac{t}{t-1} A^{-1}_{t-1} - \frac{t}{\left(t-1\right)^2} R_{t} \left(\mathrm{I}+\frac{1}{t-1} R_{t} \right)^{-1} A_{t-1}^{-1},
\end{equation*}
where
\begin{equation*}
    R_{t} := A^{-1}_{t-1} \left( \frac{\mathcal{L}_{t}^{\mathrm{T}} \nabla^2_y l\left(y_t; \zeta_t\right) \mathcal{L}_{t} }{\epsilon} + \frac{\alpha \nabla_{\omega} \pi\left(\omega_t,y_t;\zeta_t\right)   \nabla_{\omega} \pi\left(\omega_t,y_t;\zeta_t\right)^{\mathrm{T}} }{\epsilon} + \mathrm{I}\right),~\zeta_t \stackrel{\mathrm{i.i.d.}}{\sim} p_{\zeta}.
\end{equation*}
\end{proposition}

\begin{proof}
We start with the definition of $A_t$,
\begin{align*}
\begin{split}
A_{t} &= \frac{t-1}{t} A_{t-1} + \frac{1}{t} \frac{\mathcal{L}_{t}^{\mathrm{T}} \nabla^2_y l\left(y_t; \zeta_t\right) \mathcal{L}_{t} + \alpha \nabla_{\omega} \pi\left(\omega_t,y_t;\zeta_t\right) \nabla_{\omega} \pi\left(\omega_t,y_t;\zeta_t\right)^{\mathrm{T}} }{\epsilon} + \frac{1}{t}  \text{I}\\
&= \frac{t-1}{t} A_{t-1} \left( \text{I} + \frac{1}{t-1} R_t \right),~\forall  t \geq 2.
\end{split}
\end{align*}
In addition, we notice that the matrix $A_t$ is always positive definite due to the addition of the identity matrix $\mathrm{I}$. Therefore, we can calculate the inverse of $A_t$ as follows:
\begin{align*}
\begin{split}
A_{t}^{-1} =& \frac{t}{t-1} \left(\mathrm{I} + \frac{1}{t-1} R_{t}\right)^{-1} A_{t-1}^{-1}\\
=& \frac{t}{t-1} \left[\mathrm{I} - \frac{1}{t-1} R_{t} \left(\mathrm{I}+\frac{1}{t-1} R_{t} \right)^{-1} \right] A_{t-1}^{-1}\\
=&\frac{t}{t-1} A^{-1}_{t-1} - \frac{t}{\left(t-1\right)^2} R_{t} \left(\mathrm{I}+\frac{1}{t-1} R_{t} \right)^{-1} A_{t-1}^{-1},~\forall t \geq 2.
\end{split}
\end{align*}
\end{proof}
We now prove~\thmref{thm:online_quasi_newton}.
\begin{proof}
We start by analyzing the decrease of $F\left(\omega_{t+1}\right)$ conditioned on $\omega_{1:T}$, where we substitute the iterative scheme in~\algoref{algo:online_quasi_newton}. We obtain the following result:
\begin{align}
\begin{split}
 &\mathbb{E}_{\zeta_t}\left[\left. F(\omega_{t+1}) - F(\omega_t) \right| \omega_{1:t} \right]= \mathbb{E}_{\zeta_t} \left[\left. F\left(\omega_t - \eta_t A_t^{-1} \mathcal{F} \left(\omega_t;\zeta_t\right) \right) - F(\omega_t) \right| \omega_{1:t} \right]\\
\leq& \mathbb{E}_{\zeta_t}\left[\left. - \eta_t \nabla F(\omega_t)^{\text{T}} A_t^{-1} \mathcal{F} \left(\omega_t;\zeta_t \right) + \frac{L \eta_t^{2}}{2} \left|A_t^{-1} \mathcal{F} \left(\omega_t;\zeta_t\right) \right|^2 \right| \omega_{1:t}  \right]\\
\leq&  - \mathbb{E}_{\zeta_t} \left[ \left. \eta_t \nabla F(\omega_t)^{\text{T}} A_t^{-1} \mathcal{F} \left(\omega_t;\zeta_t \right) \right|\omega_{1:t} \right] + \frac{L H^2 \eta_t^{2}}{2},
\end{split}
\label{eq:first_inequality}
\end{align}
where the first inequality arises from the $L$-smoothness of $F\left(\omega\right)$, while the second inequality stems from~\aspref{asp:bounded_variance} the fact that $\nicefrac{1}{\lambda} \leq \left\|A^{-1}_t\right\| \leq 1$:
\begin{equation*}
    \mathbb{E}_{\zeta_t} \left[\left. \left|A_t^{-1} \mathcal{F} \left(\omega_t;\zeta_t \right) \right|^2 \right| \omega_{1:t} \right] \leq \left\|  A_t^{-1}\right\|^2  \mathbb{E}_{\zeta_t} \left[\left.\left|\mathcal{F} \left(\omega_t;\zeta_t \right)\right|^2 \right| \omega_{1:t} \right] \leq H^2,
\end{equation*}
where $\left\|\cdot\right\|$ denotes the spectral norm of a matrix.

Next we rearrange~\eqref{eq:first_inequality} and get the following expression:
\begin{equation}
\mathbb{E}_{\zeta_t}\left[ \left. \eta_t \nabla F(\omega_t)^{\text{T}} A_t^{-1} \mathcal{F} \left(\omega_t;\zeta_t \right) \right| \omega_{1:t} \right]
\leq \mathbb{E}_{\zeta_t}\left[ \left. F\left(\omega_{t} \right)  - F\left(\omega_{t+1} \right) \right| \omega_{1:t}\right]+\frac{L H^2 \eta_t^{2}}{2}.
\label{eq:second_inequality}
\end{equation}
By substituting the result of~\propref{prop:A_t} into the term on the left-hand side, we have the following result:
\begin{align}
\begin{split}
& \eta_t \nabla F(\omega_t)^{\text{T}} A_t^{-1} \mathcal{F} \left(\omega_t;\zeta_t \right)\\
=& \eta_t  \nabla F(\omega_t)^{\text{T}} \left( \frac{t}{t-1} A^{-1}_{t-1} - \frac{t}{(t-1)^2} R_{t} \left(\mathrm{I}+\frac{1}{t-1}R_{t}\right)^{-1} A_{t-1}^{-1}\right) \mathcal{F} \left(\omega_t;\zeta_t \right)\\
=& \frac{\eta_t t}{t-1}  \nabla F(\omega_t)^{\text{T}} A_{t-1}^{-1} \left( \nabla f \left(\omega_t;\zeta_t\right) - \nabla f \left(\omega_t;\zeta_t\right) + \mathcal{F} \left(\omega_t;\zeta_t \right) \right)\\
&- \frac{ \eta_t t}{(t-1)^2} \nabla F(\omega_t)^{\text{T}} R_{t} \left(\mathrm{I}+\frac{1}{t-1} R_{t}\right)^{-1} A_{t-1}^{-1} \mathcal{F} \left(\omega_t; \zeta_t \right)\\
\geq& \frac{ \eta_t t}{t-1}  \nabla F(\omega_t)^{\text{T}} A_{t-1}^{-1} \left( \nabla f \left(\omega_t;\zeta_t\right) - \nabla f \left(\omega_t;\zeta_t\right) + \mathcal{F} \left(\omega_t;\zeta_t \right) \right)\\
&- \frac{\eta_t \lambda}{t-1} \left| \nabla F\left(\omega_t\right) \right| \left| \mathcal{F} \left(\omega_t;\zeta_t\right) \right| ,~\forall t\geq 2.
\end{split}
\label{eq:expecation_1}
\end{align}
The last inequality is obtained due to the following facts:
\begin{equation*}
    \left\| R_t \right\| \leq \left\|A^{-1}_{t-1} \right\| \left\| \frac{\mathcal{L}_{t}^{\text{T}} 
  \nabla^2_y l \left(y_t; \zeta_t\right) \mathcal{L}_{t} }{\epsilon} + \frac{\alpha \pi_{\omega}\left(\omega_t, y_t; \zeta_t\right) \pi_{\omega}\left(\omega_t, y_t; \zeta_t\right)^{\text{T}} }{\epsilon} + \text{I} \right\| \leq \lambda,
\end{equation*}
and
\begin{multline*}
  \left\| \left(\mathrm{I}+\frac{1}{t-1}R_{t}\right)^{-1} A_{t-1}^{-1} \right\| \\
= \left\| \left(A_{t-1} + \frac{1}{t-1} \left( \frac{\mathcal{L}_{t}^{\text{T}}  \nabla^2_y l \left(y_t; \zeta_t\right)  \mathcal{L}_{t}  }{\epsilon} + \frac{\alpha \pi_{\omega}\left(\omega_t, y_t; \zeta_t\right) \pi_{\omega}\left(\omega_t, y_t; \zeta_t\right)^{\text{T}}}{\epsilon} +  \text{I} \right) \right)^{-1} \right\| \\
\leq \frac{t-1}{t}.
\end{multline*}
For $t=1$ we have
\begin{align}
\begin{split}
& \eta_1 \nabla F(\omega_1)^{\text{T}} \left( \frac{\mathcal{L}_1^{\text{T}} 
 \nabla^2_y l \left(y_1; \zeta_1\right)  \mathcal{L}_1}{\epsilon} + \frac{\alpha \pi_{\omega}\left(\omega_1, y_1; \zeta_1\right) \pi_{\omega}\left(\omega_1, y_1; \zeta_1\right)^{\text{T}}}{\epsilon} +  \text{I} \right)^{-1} \mathcal{F} \left(\omega_1;\zeta_1 \right)\\
= & \eta_1 \nabla F(\omega_1)^{\text{T}} \left( \text{I} -  R_1 \left(\text{I} + R_1\right)^{-1} \right) \mathcal{F} \left(\omega_1;\zeta_1 \right)\\
\geq & \eta_1 \nabla F(\omega_1)^{\text{T}} \left( \nabla f \left(\omega_1;\zeta_1\right) -\nabla f \left(\omega_1;\zeta_1 \right) + \mathcal{F} \left(\omega_1;\zeta_1 \right) \right)  - 
\eta_1 \lambda \left| \nabla F\left(\omega_1\right) \right| \left| \mathcal{F} \left(\omega_1;\zeta_1\right)\right|,
\end{split}
\label{eq:expecation_2}
\end{align}
where
\begin{equation*}
    R_1 := \frac{\mathcal{L}_{1}^{\mathrm{T}} \nabla^2_y l\left(y_1; \zeta_1\right) \mathcal{L}_{1} }{\epsilon} + \frac{\alpha \nabla_{\omega} \pi\left(\omega_1,y_1;\zeta_1\right)   \nabla_{\omega} \pi\left(\omega_1,y_1;\zeta_1\right)^{\mathrm{T}} }{\epsilon} + \mathrm{I},~\zeta_1 \stackrel{\mathrm{i.i.d.}}{\sim} p_{\zeta}.
\end{equation*}
We now substitute the result of~\eqref{eq:expecation_1} into~\eqref{eq:second_inequality} and rearrange the terms. We evaluate conditional expectations on both sides as follows:
\begin{align*}
\begin{split}
 &\frac{\eta_t t}{t-1} \mathbb{E}_{\zeta_t} \left[ \left. \nabla F(\omega_t)^{\text{T}} A_{t-1}^{-1}  \left( \nabla f \left(\omega_t;\zeta_t \right) -\nabla f \left(\omega_t;\zeta_t \right) + \mathcal{F} \left(\omega_t;\zeta_t \right) \right) \right| \omega_{1:t}\right]\\
 =& \frac{\eta_t t}{t-1} \nabla F(\omega_t)^{\text{T}} A_{t-1}^{-1} \nabla F\left(\omega_t \right) - \frac{\eta_t t}{t-1} \mathbb{E}_{\zeta_t} \left[ \left. \nabla F(\omega_t)^{\text{T}}  A_{t-1}^{-1} \left( \nabla f\left(\omega_t;\zeta_t \right) -  \mathcal{F} \left(\omega_t;\zeta_t \right) \right) \right| {\omega_{1:t}} \right].
\end{split}
\end{align*}
Then, we apply the Peter-Paul inequality and~\lmmref{lmm:modeling_error} to the second term, and get the following result:
\begin{align*}
    & \mathbb{E}_{\zeta_t} \left[ \left. \nabla F(\omega_t)^{\text{T}}  A_{t-1}^{-1} \left( \nabla f\left(\omega_t;\zeta_t \right) -  \mathcal{F} \left(\omega_t;\zeta_t \right) \right) \right| {\omega_{1:t}} \right]\\
=&  \nabla F(\omega_t)^{\text{T}} A_{t-1}^{-1}  \left( \nabla F\left(\omega_t\right) - \mathbb{E}_{\zeta_t}\left[ \left. \mathcal{F} \left(\omega_t;\zeta_t \right) \right| \omega_{1:t}  \right]  \right) \\
\leq& \frac{\kappa}{2} \nabla F(\omega_t)^{\text{T}} A_{t-1}^{-1} \nabla F\left(\omega_t \right) + \frac{1}{2\kappa} \left| \nabla F\left(\omega_t\right) - \mathbb{E}_{\zeta_t}\left[ \left. \mathcal{F} \left(\omega_t;\zeta_t \right) \right| \omega_{1:t}  \right]  \right|^{2}_{A_{t-1}^{-1}} \\
\leq &\frac{\kappa}{2} \left| \nabla F(\omega_t)\right|^2_{A_{t-1}^{-1}} + \frac{\kappa}{2} \left| \nabla F(\omega_t)\right|^2_{A_{t-1}^{-1}} \\
=& \kappa \left| \nabla F(\omega_t)\right|^2_{A_{t-1}^{-1}}.
\end{align*}
Meanwhile, we have
\begin{equation*}
    \mathbb{E}_{\zeta_t} \left[\left. \left| \nabla F\left(\omega_t\right) \right| \left| \mathcal{F} \left(\omega_t;\zeta_t\right) \right| \right| \omega_{1:t}\right] \leq \mathbb{E}_{\zeta_t} \left[\left. \left| \nabla F\left(\omega_t\right) \right| \right| \omega_{1:t}\right] \mathbb{E}_{\zeta_t} \left[\left. \left| \mathcal{F} \left(\omega_t;\zeta_t\right) \right| \right| \omega_{1:t}\right] \leq H^2,
\end{equation*}
as a result of Jensen's inequality. At last, we have
\begin{align*}
    \frac{\eta_t t}{t-1} \mathbb{E}_{\zeta_t} \left[ \left. \nabla F(\omega_t)^{\text{T}} A_{t-1}^{-1}  \mathcal{F} \left(\omega_t;\zeta_t \right) \right| \omega_{1:t}\right] \geq \frac{\eta_t t \left(1-\kappa\right)}{t-1} \left|\nabla F\left(\omega_t\right) \right|^2_{A_{t-1}^{-1}} - \frac{\eta_t H^2 \lambda}{t-1}.
\end{align*}
For $t=1$ we have
\begin{equation*}
\eta_1 \mathbb{E}_{\zeta_1} \left[ \left.  \nabla F(\omega_1)^{\text{T}} \mathcal{F} \left(\omega_1;\zeta_1 \right) \right| \omega_1 \right]
\geq \eta_1 \left(1-\kappa\right) \left| \nabla F(\omega_1) \right|^2 - \eta_1 H^2 \lambda.
\end{equation*}

Further, we get the following result by considering the expectation over all random variables on both sides in~\eqref{eq:second_inequality}:
\begin{multline}
\left(1-\kappa\right)\frac{\eta_t t}{t-1} \mathbb{E}_{\zeta_{1:T}}\left[ \left| \nabla F(\omega_t) \right|^2_{A^{-1}_{t-1}} \right] \\
\leq \mathbb{E}_{\zeta_{1:T}}\left[F\left(\omega_{t} \right)\right]  - \mathbb{E}_{\zeta_{1:T}}\left[F\left(\omega_{t+1} \right) \right]
+ \frac{L H^2 \eta_t^{2}}{2} + \frac{\eta_t H^2 \lambda}{t-1},~\forall t \geq 2.
\label{eq:result_1}
\end{multline}
We note that
\begin{equation*}
\mathbb{E}_{\zeta_{1:T}}\left[ \left| \nabla F(\omega_t) \right|^2_{ A_{t}^{-1}}\right] \leq \frac{t}{t-1} \mathbb{E}_{\zeta_{1:T}}\left[ \left| \nabla F(\omega_t) \right|^2_{ A_{t-1}^{-1}} \right],~\forall t \geq 2,
\end{equation*}
and
\begin{equation*}
\mathbb{E}_{\zeta_{1:T}}\left[ \left| \nabla F(\omega_1) \right|^2_{A_{1}^{-1}} \right] \leq \mathbb{E}_{\zeta_{1:T}}\left[ \left| \nabla F(\omega_1) \right|^2\right].
\end{equation*}
In the following, we exploit the fact that the step size is constant, that is, $\eta_t=\eta$ for all $t \geq 1$. As a consequence, \eqref{eq:result_1} can be rewritten as follows:
\begin{multline*}
\left(1-\kappa\right)\mathbb{E}_{\zeta_{1:T}}\left[\left| \nabla F(\omega_t) \right|^2_{A_{t}^{-1}} \right] \\
\leq \frac{ \mathbb{E}_{\zeta_{1:T}}\left[F(\omega_t)\right] -  \mathbb{E}_{\zeta_{1:T}}\left[F(\omega_{t+1})\right]}{\eta} + \frac{L H^2 \eta}{2} + \frac{H^2 \lambda}{t-1},~\forall t \geq 2.
\end{multline*}
By summing up the above equation and including the case $t=1$ we get the following result:
\begin{multline*}
\left(1-\kappa \right)\sum_{t=1}^{T}  \mathbb{E}_{\zeta_{1:T}}\left[\left| \nabla F(\omega_t) \right|^2_{A_{t}^{-1}}\right]\\
\leq \frac{ F\left(\omega_{1} \right) -  F\left(\omega^{\star} \right)}{\eta} + \frac{ TLH^2 \eta}{2} +  H^2 \lambda\sum_{t=1}^{T}a_t,~\forall t \geq 1,
\end{multline*}
where we notice that $F(\omega^{\star}) \leq \mathbb{E}_{\zeta_{1:T}} \left[F(\omega_{T+1})\right]$ always holds since $\omega^{\star}$ denotes the global minimum, and $a_t$ is defined as $a_1 = 1$, $a_t = \nicefrac{1}{t-1}$ for $t\geq2$. The above equation can be further simplified due to the fact that $F\left(\omega^{\star}\right) \geq 0$:
\begin{equation}
\left(1-\kappa \right)\sum_{t=1}^{T}  \mathbb{E}_{\zeta_{1:T}}\left[\left| \nabla F(\omega_t) \right|^2_{A_{t}^{-1}}\right] \leq \frac{ F\left(\omega_{1} \right)}{\eta} + \frac{ TLH^2 \eta}{2} +  H^2 \lambda\sum_{t=1}^{T}a_t,~\forall t \geq 1.
\label{eq:result_2}
\end{equation}
The sum over $a_t$ is bounded as follows:
\begin{align}
\begin{split}
\sum_{t=1}^{T}a_t &\leq \int_{2}^{T} \frac{1}{t-1} \mathrm{d}t + 2\\
&= \ln{\left(T-1\right)} + 2\\
&\leq \ln{T} + 2,~\forall T \geq 1.
\end{split}
\label{eq:int_a}
\end{align}
We substitute~\eqref{eq:int_a} into~\eqref{eq:result_2} and calculate its average value as follows:
\begin{equation}
\left(1-\kappa\right) \frac{1}{T} \sum_{t=1}^{T}  \mathbb{E}_{\zeta_{1:T}} \left[ \left| \nabla F(\omega_t) \right|^2_{A_{t}^{-1}} \right] \leq \frac{ F\left(\omega_{1} \right)}{\eta T} + \frac{LH^2 \eta}{2} + \frac{\lambda H^2 \left(\ln{T} + 2\right)}{T}.
\end{equation}
We set the first two terms on the right-hand side to be equal by choosing the step size appropriately,
\begin{equation*}
\frac{ F\left(\omega_{1} \right)}{\eta T} =  \frac{ LH^2 \eta}{2}
\Leftrightarrow  \eta = \sqrt{\frac{2 F\left(\omega_{1} \right) }{LH^2 T}}.
\end{equation*}
This yields the following result:
\begin{equation*}
 \frac{1-\kappa}{T} \sum_{t=1}^{T}  \mathbb{E}_{\zeta_{1:T}}\left[\left| \nabla F(\omega_t) \right|^2_{A_{t}^{-1}} \right]\leq \sqrt{\frac{2 LH^2 F\left(\omega_{1}  \right)}{T}}  + \frac{\lambda H^2 \left(\ln{T} + 2\right)}{T},
\end{equation*}
which is dominated by the first term on the right-hand side of the inequality for large $T$. It also implies that the average expected gradient will converge to zero at a rate of $\mathcal{O}\left(\nicefrac{1}{\sqrt{T}}\right)$.
\end{proof}

\section{Derivation of Trust-Region Approach}
\label{sec:app_trust_region}
In this appendix, we will present in detail the necessary intermediate steps used to derive the results in \secref{sec:interpretation}. We start from the local approximation of the function $M\left(v, \omega\right)$
\begin{multline}
M \left(v, \omega \right) = \underbrace{f \left(\omega \right) + \nabla f \left(\omega\right)^{\text{T}}  \left(v-\omega\right)+\frac{1}{2} \left(v-\omega\right)^{\text{T}} \nabla^2 f\left(\omega\right) \left(v-\omega\right)}_{\text{second order Taylor expansion}}\\
+\underbrace{\frac{\alpha}{2} \left|\pi \left( v \right) - \pi \left(\omega \right) \right|^2}_{\text{Term 1}} +\underbrace{\frac{\epsilon}{2} \left|v-\omega\right|^2}_{\text{Term 2}}.
\label{eq:trust_region}
\end{multline}
To obtain~\eqref{eq:intermediate_trust_region}, it is important to note that, as specified in~\eqref{eq:trust_region}, calculating the gradient and Hessian matrix of the loss function $f$ is essential, which can be expressed as follows:
\begin{align}
    \begin{split}
        \nabla f\left(\omega\right) &= \frac{\partial y^{\text{T}}}{\partial \omega} \nabla_y l\left(y;\zeta\right),\\
        \nabla^2 f\left(\omega\right) &=  \frac{\partial^2 y^{\text{T}}}{\partial \omega^2}  \nabla_y l\left(y;\zeta\right) + \frac{\partial y^{\text{T}}}{\partial \omega}  \nabla^2_y l\left(y;\zeta\right) \frac{\partial y }{\partial \omega},
    \end{split} 
\end{align}
where we introduce
\begin{equation*}
    \frac{\partial y}{\partial \omega} \approx \mathcal{L}= \left(\mathrm{I} - \mathcal{G}\left(u\right)\frac{\partial \pi\left(\omega,y;\zeta\right)}{\partial y}\right)^{\dagger} \mathcal{G} \left(u\right) \frac{\partial \pi \left(\omega,y;\zeta\right)}{\partial \omega},
\end{equation*}
and assume
\begin{equation*}
    \frac{\partial^2 y}{\partial \omega^2} \approx 0.
\end{equation*}
Then, \eqref{eq:trust_region} can be formulated as follows:
\begin{multline*}
M\left(v, \omega \right) = f \left(\omega\right) + \left(v-\omega \right)^{\text{T}} \mathcal{L}^{\text{T}} \nabla_y l\left(y;\zeta\right)  +\frac{1}{2} \left(v-\omega\right)^{\text{T}} \mathcal{L}^{\text{T}} \nabla^2_y l\left(y;\zeta\right) \mathcal{L} \left(v-\omega\right)\\
+\frac{\alpha}{2} \left|\pi \left( v \right) - \pi \left(\omega \right) \right|^2 + \frac{\epsilon}{2} \left|v-\omega\right|^2.
\end{multline*}
We further perform a Taylor expansion of $\pi\left(v\right)$ around $\pi\left(\omega\right)$, that is, 
\begin{equation*}
    \pi\left(v\right) = \pi\left(\omega\right) + \left.\frac{\partial \pi\left(\omega\right)}{\partial \omega}\right|_{\omega} \left(v - \omega\right) + \text{HOT},
\end{equation*}
where HOT denotes the remaining higher order term. This yields~\eqref{eq:intermediate_trust_region}.

To obtain~\eqref{eq:closed_form_solution}, we calculate the expectation of $M\left(v, \omega\right)$ as follows:
\begin{multline}
\mathbb{E}_{\zeta} \left[M(v, \omega)|\omega\right] = \underbrace{\int f\left(\omega;\xi\right) p_{\zeta} \left(\xi\right) \text{d} \xi}_{F\left(\omega\right)} + \left(v-\omega\right)^{\text{T}} \int \mathcal{L} \left(\xi\right)^{\text{T}} \nabla_y l\left(y; \xi\right) p_{\zeta} \left(\xi\right) \text{d} \xi \\
+ \frac{1}{2} \left(v-\omega\right)^{\text{T}}\underbrace{\int  \mathcal{L} \left(\xi\right)^{\text{T}} \nabla^2_y l\left(y; \xi\right) \mathcal{L} \left(\xi\right) p_{\zeta}\left(\xi\right) \text{d} \xi}_{\epsilon{\Sigma}_{1}} \left(v-\omega\right)\\
+ \frac{\alpha}{2} \left(v-\omega\right)^{\text{T}} \underbrace{\int \nabla_{\omega} \pi\left(\omega;\xi\right) \nabla_{\omega} \pi \left(\omega;\xi\right)^{\text{T}}~p_{\zeta} \left(\xi\right) \text{d} \xi}_{\epsilon {\Sigma}_{2}} \left(v-\omega\right)\\
+\frac{\epsilon}{2} \left(v-\omega\right)^{\text{T}} \left(v-\omega\right).
\end{multline}

\section{Pretraining Process}
\label{sec:AppendixB}
\begin{figure}[H]
\centering
\begin{subfigure}{0.45\textwidth}
    \centering
    \includegraphics[width=\linewidth]{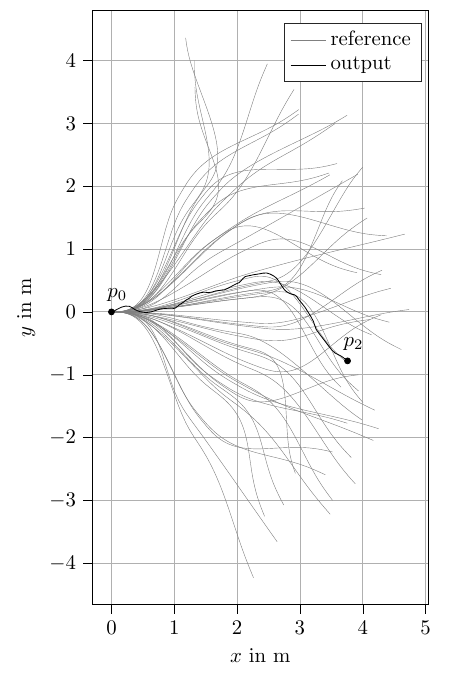}
    \caption{distribution of trajectories for ILC}
\end{subfigure}
\hskip 0cm
\begin{subfigure}{0.50\textwidth}
    \centering
    \includegraphics[width=\linewidth]{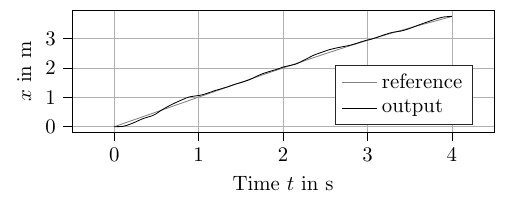}
    \caption{trajectory of $x$}
    \includegraphics[width=\linewidth]{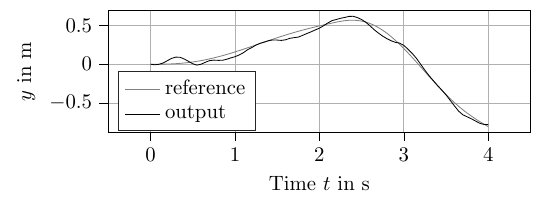}
    \caption{trajectory of $y$}
    \includegraphics[width=\linewidth]{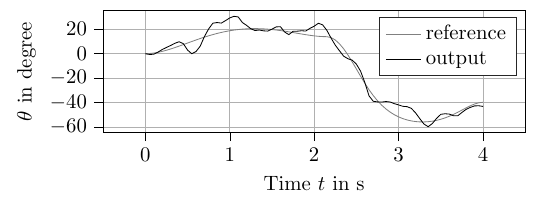}
    \caption{trajectory of yaw}
\end{subfigure}
\caption{The left subfigure shows the distribution of $\num{50}$ randomly sampled reference trajectories used for pre-training (represented by gray lines), and illustrates the tracking performance of ILC with one of these trajectories (depicted as a black line). The right subfigures demonstrate the tracking performance of the ILC for the particular trajectory in the $x$, $y$, and yaw components.}
\label{fig:ant_ilc}
\end{figure}

In this appendix, we will briefly explain the pretraining process that is used in~\secref{sec:ant}. The matrices $U$ and $V$ are obtained through the singular value decomposition of the matrix $R$: 
\begin{equation*}
    R = U \mathrm{diag} \left(\sigma\right) V^{\mathrm{T}},
\end{equation*}
where the matrix $R$ is derived by solving the following ridge regression:
\begin{equation*}
    \min_{R \in \mathbb{R}^{mq \times nq}}~\frac{1}{2} \sum_{i=1}^{n_{\text{ILC}}}\left|u_{\text{ref}, i} -  R y_{\text{ref}, i}   \right|^2 + \frac{\rho}{2} \left\| R\right\|^2_{\text{F}},
\end{equation*}
where $\rho$ is a positive constant, and $\left\|\cdot\right\|_{\text{F}}$ denotes the Frobenius norm. The ideal input $u_{\text{ref}}$ represents the input required for accurately tracking a given reference trajectory $y_{\text{ref}}$. The ideal input is unknown, and we therefore employ iterative learning control (ILC) to approximate it~\citep{maLearningbasedIterativeControl2022b, hoferIterativeLearningControl2019, zughaibiFastReliablePickandPlace2021, muellerIterativeLearningFeedforward2012, schoelligOptimizationbasedIterativeLearning2012,zughaibiBalancing3DInverted2024}. The variable $n_{\text{ILC}}$ denotes the number of pre-trained trajectories using ILC. In this experiment, we sample $\num{50}$ reference trajectories and get their corresponding ideal inputs using ILC, and each reference trajectory takes $\num{200}$ to $\num{300}$ iterations to obtain the ideal inputs. We then use $\num{45}$ trajectories ($\num{90}\%$) along with their ideal inputs to perform ridge regression. \figref{fig:ant_ilc} displays the distribution composed of all $\num{50}$ reference trajectories used for pre-training in the left subfigure, whereas the right subfigure showcases the final training result of ILC for one reference trajectory as an example. The right subfigures illustrate the tracking performance of the corresponding $x$, $y$, and yaw components of this trajectory. We note that the tracking of the $x$ and $y$ components by the ILC is very effective; however, due to the presence of collisions, the tracking of the yaw component is slightly less accurate. Nevertheless, we consider this as a sufficiently good ideal input for tracking the given reference trajectory, which is able to capture the motion patterns.

\end{appendices}


\bibliography{sn-bibliography}
\end{document}